\documentclass[fleqn,10pt]{wlscirep}
\usepackage[utf8]{inputenc}
\usepackage[T1]{fontenc}
\usepackage{bm}
\usepackage{hyperref}       
\usepackage{url}            
\usepackage{booktabs}       
\usepackage{amsfonts}       
\usepackage{nicefrac}       
\usepackage{amsmath}
\usepackage{microtype}      
\usepackage{lipsum}
\usepackage{graphicx}
\usepackage{cleveref}
\crefrangelabelformat{equation}{#3#1#4--#5#2#6}
\usepackage{tikz}

\usepackage{geometry}
\usepackage{booktabs,tabularx}
\usepackage{rotating}
\usepackage{makecell}
\usepackage{pifont}
\usepackage{array}
\usepackage[breakable]{tcolorbox}
\usepackage{longtable}

 \tcbset{
 	mybox/.style={
 		colback=gray!5!white,  
 		colframe=black!20, 
 		fonttitle=\bfseries,    
 		coltitle=black,         
 		boxrule=0.5mm,         
 		arc=2mm,                
 		left=4mm,               
 		right=4mm,              
 		top=2.5mm,                
 		bottom=2mm,             
 		width=\textwidth,       
 		before=\medskip,        
 		after=\medskip,         
 		breakable
 	}
 }
\title{Data-driven discovery of governing differential equations across physical systems}

\author[1,2]{Siyu Lou}
\author[2]{Hao Xu}
\author[3]{Wenguan Wang}
\author[4,5]{Lu Lu}
\author[6]{Hao Sun}
\author[7]{Yang Liu}
\author[8]{Linfeng Zhang}
\author[2]{Dongxiao Zhang}
\author[2,*]{Yuntian Chen}
\affil[1]{School of Computer Science, Shanghai Jiao Tong University, Shanghai, China }
\affil[2]{Ningbo Key Laboratory of Advanced Manufacturing Simulation, Eastern Institute of Technology, Ningbo, Ningbo, China }
\affil[3]{The State Key Lab of Brain-Machine Intelligence, Zhejiang University, Hangzhou, China.}
\affil[4]{Department of Statistics and Data Science, Yale University, New Haven, CT, USA}
\affil[5]{Department of Chemical and Environmental Engineering, Yale University, New Haven, CT, USA}
\affil[6]{Gaoling School of Artificial Intelligence, Renmin University of China, Beijing, China}
\affil[7]{School of Engineering Sciences, University of Chinese Academy of Sciences, Beijing, China}
\affil[8]{DP Technology, Beijing, China}

\affil[*]{corresponding author, e-mail: ychen@eitech.edu.cn}

\begin{abstract}

Differential equations play a critical role in scientific discovery because they provide a mathematical framework to describe the behaviour of physical phenomena. As a promising alternative to traditional first principles, data-driven differential equation discovery has attracted increasing attention for its ability to infer governing laws directly from experimental or simulated data, especially when the underlying physics is unclear. However, the field has expanded rapidly along diverse methodological directions, particularly with the emergence of AI-based approaches, and still lacks a clear organizing perspective.
In this Review, we propose a problem-oriented perspective on data-driven differential equation discovery. We first introduce a two-dimensional phase diagram of equation discoverability, where discovery problems are organized according to structural complexity and coefficient complexity. This phase diagram shows how the field has moved from the discovery of sparse equations with simple coefficients toward more complex governing laws with richer structures and more flexible parameterizations. It also clarifies why different methodological families succeed or fail in different problem settings. We then present the representation–evaluation–optimization (REO) framework as a fundamental abstraction of the discovery process. By identifying the core problems of equation discovery that persist across algorithmic variations, REO shifts the discussion from individual algorithms to the fundamental principles that determine discoverability. We connect these perspectives to applications across physics and adjacent sciences, and argue that the next challenge is not merely recovering equations, but using them to revise existing theories, distil mechanisms and form new scientific concepts.

\end{abstract}
\begin{document}

\flushbottom
\maketitle

\thispagestyle{empty}


\section*{Introduction}

Equations play an important role in scientific advancement because they transform empirical observations into compact symbolic forms that can be communicated among researchers and used for deductive reasoning in new settings~\cite{popper2005logic}. In particular, differential equations provide a natural language for formalizing dynamic systems. By expressing these processes as derivatives in time and/or space, differential equations offer a concise and expressive framework for capturing governing laws across a wide range of systems, such as the Navier--Stokes equations for fluid motion~\cite{batchelor2000introduction}, the Gray--Scott equations for chemical pattern formation~\cite{pearson1993complex}, gravitational $N$-body equations for celestial dynamics~\cite{binney2011galactic}, and constitutive equations in continuum mechanics for stress--strain relations with material memory~\cite{holzapfel2000nonlinear}.

\begin{figure}
	\centering
	\includegraphics[width=0.95\linewidth]{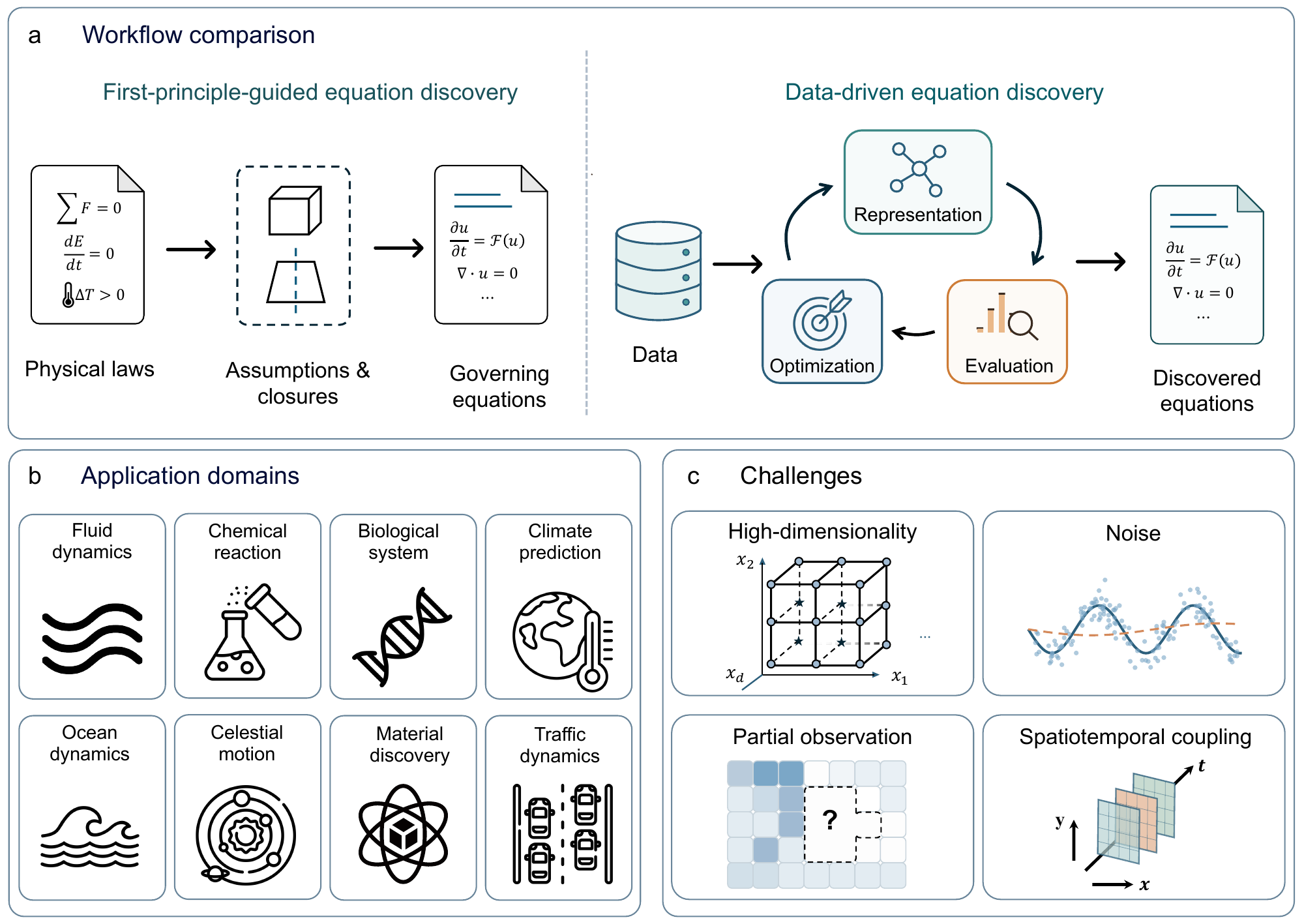}
    \caption{\textbf{Overview of data-driven differential equation discovery.} \textbf{a} Comparison between traditional first-principles and data-driven workflows for discovering differential equations. \textbf{b} Representative application domains in which differential equation discovery is used to model complex systems and uncover governing dynamics. \textbf{c} Major challenges in data-driven differential equation discovery.}
	\label{fig:overview}
\end{figure}

Abstracting observations into explicit equations remains a major challenge. Classical scientific modelling typically proceeds from first principles, as illustrated in \Cref{fig:overview}a. This succeeds when the systems are sufficiently simple. However, many real-world systems are high-dimensional and complex (such as strongly nonlinear and multiscale), and intuition alone struggles to uncover governing equations. For example, climate dynamics involve interacting processes across the atmosphere, ocean, land, and ice, generating coupled dependencies that are difficult to compress into transparent mathematical relations by manual inspection. At the same time, modern machine-learning methods can accurately fit highly complex dynamical behaviour~\cite{dong2025recent,pathak2018model,lu2021learning,lifourier2021}. However, these models often operate as input--output approximators, lacking transparency and interpretability.

Data-driven differential equation discovery, as illustrated in \Cref{fig:overview}a, has emerged as a promising approach for recovering explicit and interpretable governing laws directly from data. Early efforts often focused on fitting simple parametric forms, such as linear models or polynomial expansions, under relatively strong assumptions about the candidate equation class. Subsequent work using genetic algorithms for nonlinear system identification showed that both the structure and parameters of governing equations could, in principle, be inferred from data~\cite{bongard2007automated,schmidt2009distilling}. Sparse regression methods, most notably sparse identification of nonlinear dynamical systems (SINDy)~\cite{brunton2016discovering}, then provided a more scalable route to parsimonious and interpretable model discovery from observations. Although much of the early literature centered on ordinary differential equations, recent studies have increasingly turned to partial differential equation discovery, where spatial coupling, derivative estimation, and limited observability introduce additional challenges.

This methodological progress has been accompanied by a growing range of scientific applications. Equation discovery has been used in settings spanning space physics, active-matter hydrodynamics, biological dynamics, turbulent-flow modelling, geoscience, and chemical processes~\cite{course2023state,supekar2023learning,reinbold2021robust,li2025bi,beetham2020formulating,beetham2021sparse,song2025deep,zhan2024physics,xu2025explicit,ying2025neural}, with additional domains summarized in \Cref{fig:overview}b. These examples suggest that equation discovery is evolving from a primarily methodological enterprise into a practical framework for extracting compact and physically meaningful descriptions of complex systems.

At the same time, differential equation discovery poses challenges that distinguish it from the discovery of algebraic relations, which has a longer history. Governing equations often involve spatiotemporal coupling, differentiation of noisy data, hidden variables, boundary effects, and partial observability (\Cref{fig:overview}c). As a result, the field now encompasses a diverse set of approaches that differ not only in their algorithms, but also in the assumptions they encode, the data regimes they support, and the forms of scientific knowledge they can extract. While this diversity has accelerated progress, it also makes the rapidly expanding literature increasingly difficult to navigate.

Despite this rapid growth, most existing reviews organize the field by algorithmic family or application domain~\cite{north2023review, brunton2024promising, gennemark2009benchmarks, shojaeellm, sanderse2025scientific}. These summaries are valuable, but a method- or domain-centered perspective can obscure the deeper question driving the field: which equations can actually be discovered from the available data, and what makes them scientifically meaningful? Without a framework that connects the structure of the target equation, the chosen representation, and the evaluation of scientific value, it remains difficult to compare methods or to understand when a discovered equation can truly advance knowledge.

To address this gap, we depart from a purely method-centred organization. We first develop a two-dimensional phase diagram of equation discoverability, in which discovery problems are positioned according to structural complexity and coefficient complexity. This perspective provides a problem-oriented map of the field, revealing distinct regimes of inference and the corresponding demands they place on data, prior knowledge, and computational strategy. We then introduce the representation--evaluation--optimization (REO) framework as an abstract decomposition of the equation discovery task itself. Rather than serving as a taxonomy of methods, the REO framework identifies three fundamental components of discovery: representation of candidate equations, evaluation of their validity, and optimization of the search process. These two perspectives provide a stable conceptual basis for navigating a rapidly evolving field in which new methods continue to emerge across algorithms and applications.

Finally, we review representative applications across scientific disciplines to show how differential equation discovery is being used in real-world settings. We emphasize, however, that equation recovery is not an end in itself. The broader promise of the field is conceptual: discovered equations can revise existing concepts, make implicit modelling knowledge explicit, and support closed-loop discovery processes that generate new concepts. We conclude by framing these three routes as a roadmap for the next stage of differential equation discovery, in which data-driven methods not only recover governing laws but also help extend the theoretical language of science.

\section*{Formulation of differential equation discovery}\label{sec:2}

Equation discovery aims to identify the mathematical relations governing a system directly from observed data. These relations may take the form of algebraic, stochastic, or differential equations. In this review, we focus on differential equation discovery, since differential equations provide a fundamental framework for describing the evolution of physical, biological, and chemical systems in time and/or space.

A scalar differential equation (\(m=1\)) or a coupled system of differential equations (\(m>1\)) can be written in the general implicit operator form
\begin{equation}\label{eq:de}
    \mathcal{F}(\mathbf{z}, \mathbf{u}, \nabla \mathbf{u}, \nabla^2 \mathbf{u}, \ldots)=0,
\end{equation}
where \(\mathbf{z}=(t,\mathbf{s})\in\Omega\subset\mathbb{R}^{d+1}\) denotes the independent variables, consisting of time \(t\) and spatial coordinates \(\mathbf{s}\), and \(\mathbf{u}(\mathbf{z})\in\mathbb{R}^m\) denotes the system state. When \(\mathbf{u}\) depends on a single independent variable, such as time \(t\), this formulation reduces to an ordinary differential equation (ODE). When it depends on multiple independent variables, such as time and space, it represents a partial differential equation (PDE). Differential equation discovery then seeks to infer the functional form of the operator \(\mathcal{F}\) from discrete observations \(\{\mathbf{z}^i,\mathbf{u}(\mathbf{z}^i)\}_{i=1}^{N}\), where \(N\) is the number of observation points.

In many practical discovery methods, particularly in sparse-regression and symbolic discovery methods, the governing relation is further written in evolution form,
\begin{equation}\label{eq:de-evolution}
\mathbf{u}_t = \sum_{k} \theta_k\, \Phi_k\big(\mathbf{u},\nabla \mathbf{u},\nabla^2 \mathbf{u},\ldots\big),
\end{equation}
where the functions \(\Phi_k\) denote candidate structural components and the coefficients \(\theta_k\) specify their associated weights. This representation makes explicit two major sources of difficulty in equation discovery: the complexity of the structural hypothesis space \(\{\Phi_k\}\) and the complexity of the coefficient space \(\{\theta_k\}\).

From an inference perspective, differential equation discovery can be viewed as an inverse problem in which both the form and parameters of the governing equations must be inferred from observed data. In general, this problem is highly ill-posed: distinct governing equations may yield trajectories that are observationally indistinguishable over finite time intervals, especially when measurements are noisy or incomplete. Such non-uniqueness is one of the central challenges in data-driven equation discovery.

\section*{A phase diagram of equation discoverability}\label{sec3}
\begin{figure}
	\centering
	\includegraphics[width=\linewidth]{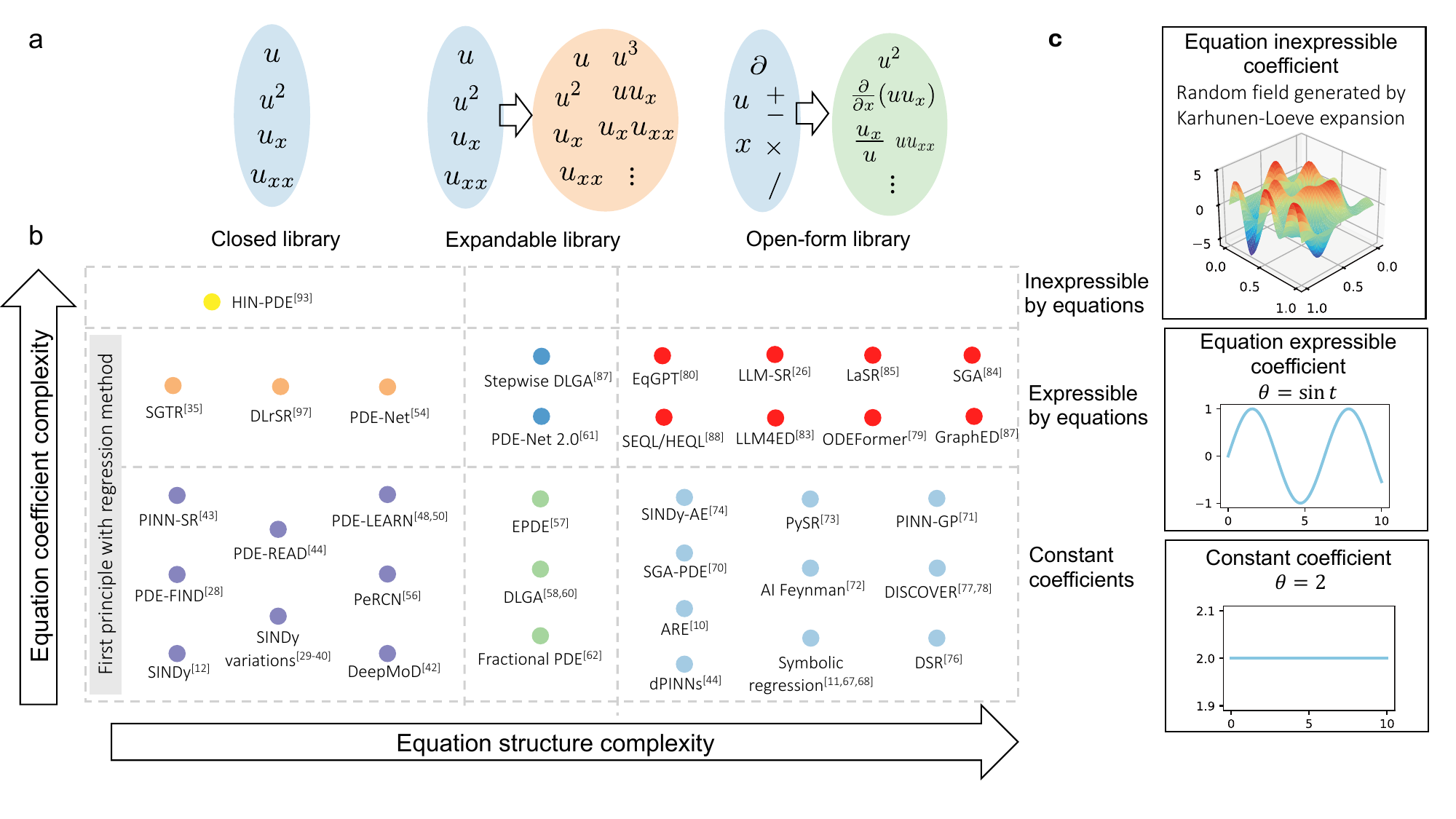}
	\caption{\textbf{A phase diagram of equation discoverability based on structure and coefficient complexity.} \textbf{a} 
    Illustration of closed library, expandable library, and open-form library. \textbf{b} A two-dimensional taxonomy of existing differential equation discovery algorithms. The horizontal axis indicates increasing equation structure complexity (from closed-form libraries to open-form libraries), while the vertical axis indicates increasing coefficient complexity (from constant coefficients to coefficients that are inexpressible by explicit equations). Each point corresponds to a published method, and the nine dashed regions form a conceptual $3\times3$ grid representing different combinations of structure and coefficient complexity. Points' relative positions within each grid cell do not imply quantitative performance comparisons. \textbf{c} Examples of three types of coefficients: constant coefficients, coefficients expressible by equations, and coefficients that are inexpressible by equations.}
	\label{fig:category}
\end{figure}

Differential equation discovery is governed by two coupled inferential tasks: discovering the functional structure of the equation, \emph{i.e.}, the explicit functional forms $\{\Phi_k\}$ in \Cref{eq:de-evolution}, and determining its associated coefficients, \emph{i.e.}, $\{\theta_k\}$ in \Cref{eq:de-evolution}.
We therefore introduce a two-dimensional phase diagram of equation discoverability. Along the horizontal axis, \emph{structural complexity} increases from closed-form libraries of predefined terms to open-form symbolic or generative representations. Along the vertical axis, \emph{coefficient complexity} increases from constant parameters to coefficients that vary across space, time, or stochastic realizations, and in the most difficult cases may not be expressible through explicit functional relations. Different regions of this space correspond to qualitatively distinct regimes of inference, with different requirements on data quality, prior knowledge, and computational strategy (\Cref{fig:category}).

Importantly, this landscape does more than summarize existing methods. It shows how the fundamental difficulty of scientific inference changes across regimes. In the low-left regime, where both equation structure and coefficients are simple, discovery is often well-posed and can be addressed effectively with library-based sparse regression. As structural complexity or coefficient complexity increases, the problem becomes progressively more ill-posed, requiring more expressive representations, stronger physical constraints, and more powerful optimization strategies. The upper-right regime, which combines open-form structure with highly complex or inexpressible coefficients, marks the current frontier of equation discovery and, to our knowledge, remains underexplored. It therefore defines a central challenge for future work.

\subsection*{Discovering differential equations with increasing structural complexity}
The expressive power of differential equations arises largely from the functional form of individual operator terms. In systems governed by a small number of low-order interactions, the dynamics may be well approximated within a closed functional basis, allowing governing relations to be recovered using library-based discovery methods. However, as system dynamics become increasingly nonlinear or involve nested functional dependencies, the admissible hypothesis space expands beyond any predefined dictionary of candidate terms.

Closed-library approaches assume that the governing relation lies within a finite functional basis constructed from combinations of state variables and their derivatives. While this assumption enables efficient inference through sparse regression, it restricts the set of discoverable operators to those expressible within the chosen basis. Consequently, discovery may fail when the true dynamics involve non-polynomial, rational, or composite functional forms.
Expandable-library methods relax this assumption by generating additional candidate terms through symbolic operations or evolutionary rules. Although these approaches increase expressiveness, the resulting hypothesis space remains constrained by the underlying generative rules used to construct new terms.
Open-form discovery methods remove these restrictions by representing governing equations as compositional symbolic structures. In principle, such approaches can express arbitrary mathematically valid operators; however, the resulting hypothesis space grows combinatorially with equation complexity, introducing substantial challenges for structural search and identifiability.

\subsubsection*{Closed-library methods}
Closed-library methods usually contain a predefined dictionary of candidate terms, which typically include combinations of the system's state variables and their spatial or temporal derivatives, often expressed as polynomials up to a specified order. This line of research began with SINDy~\cite{brunton2016discovering} and later extended to the PDE variant PDE-FIND~\cite{rudy2017data}. In this direction, various methods have been proposed to address noisy and sparse data. WeakSINDy~\cite{messenger2021weak1} and its PDE variant~\cite{messenger2021weak2} leverage the weak formulation of the dynamics to replace local pointwise derivative approximations. Variational framework and statistical test for PDE discovery are proposed to improve noise tolerance and stability of sparse regression-based methods\cite{wang2019variational,wang2021variational}.
To ensure physically consistent results, constrained SINDy~\cite{loiseau2018constrained, kaiser2018sparse} incorporates physics-informed constraints, such as symmetries, invariants,  dimensional consistency, and sign rules, to eliminate spurious terms. Group-sparse~\cite{rudy2019data} and ensemble-based methods~\cite{fasel2022ensemble}, including stability-selection variants\cite{maddu2022stability}, use multiple trajectories and resampling techniques to disambiguate correlated features and quantify structural uncertainty. Variable-coefficient and parametric extensions further improve the flexibility of PDE discovery by augmenting the candidate library with basis functions over space or parameters, allowing the capture of spatial and parametric heterogeneity, as explored by Schaeffer\cite{schaeffer2017learning} and Rudy et al.\cite{rudy2019data}. Derivative-free approaches, such as integral SINDy, mitigate the challenges posed by noisy derivative measurements and can be combined with weak-form PDEs to improve robustness\cite{kaheman2020sindy}. Finally, the practical implementation of these advancements has been greatly facilitated by PySINDy, a unified software library that integrates techniques like STLSQ/STRidge, LASSO, ensembling, implicit/rational extensions (SINDy-PI), and constraint handling into a user-friendly framework\cite{kaptanoglu2022pysindy}.

One critical challenge in discovering governing equations from data in dynamic systems is the accurate calculation of derivatives. 
Finite difference methods often become unreliable when data are sparse, noisy, or coarsely resolved, particularly for higher-order derivatives; related limitations of conventional discretizations have also motivated learned numerical representations of PDE operators~\cite{bar2019learning}. 
To address this issue, the development of neural networks with automatic differentiation has proven highly effective. DeepMoD~\cite{both2021deepmod} employs a neural network to learn the mapping from the data, \emph{e.g.}, $(\mathbf{z}, t) \mapsto \mathbf{u}$. During training, in addition to adjusting the weights and biases of the network, it also optimizes the coefficients $\{\theta_k\}$ in \Cref{eq:de-evolution} to be sparse. 
Similarly, PINN-SR~\cite{chen2021physics} adopts a simple but effective optimization strategy, \emph{i.e.}, alternating the training of the neural network's trainable parameters and sparse PDE coefficients. PINN-SR demonstrates strong efficacy and robustness across various PDEs under different levels of data scarcity and noise, consistent with other noise-aware frameworks~\cite{thanasutives2023noise, yuan2023machine}. PDE-READ~\cite{stephany2022pde} utilizes two rational neural networks~\cite{boulle2020rational}, with trainable activation functions, thereby enhancing the neural network's approximation capability. Building on PDE-READ and DeepMoD, PDE-LEARN~\cite{stephany2024pde} enhances the loss function by incorporating elements of the iteratively reweighted least-squares algorithm~\cite {chartrand2008iteratively} to promote sparsity in the coefficients. Additionally, PDE-LEARN employs an adaptive procedure to place extra points in regions where the loss is highest. These improvements accelerate convergence and result in an algorithm that is highly effective in scenarios with limited and noisy data. Afterward, Weak-PDE-LEARN is proposed to employ a weak-form variation to further enhance robustness in noisy data scenarios~\cite{stephany2024weak}. In parallel, Kolmogorov–Arnold Networks (KANs)~\cite{liu2025kan} have also been explored for PDE discovery, with KAN-ODEs~\cite{koenig2024kan} and optPDE~\cite{kantamneni2024optpde} demonstrating the ability to learn hidden dynamics and identify ODE/PDE forms from sparse or noisy datasets.

Another prominent approach, PDE-Net, learns differential operators by training convolution kernels (filters) through neural networks~\cite{long2018pde}. To further enhance noise robustness, differential spectral normalization has been integrated with PDE-Net, improving its predictive power in challenging scenarios~\cite{so2021differential}. Additionally, Rao et al.\cite{rao2022discovering} propose a deep convolutional-recurrent network that encodes prior physics knowledge, such as known PDE terms or assumed structures, to effectively handle low-quality data. 

However, these methods that rely on a predefined library of candidate terms often face limitations in flexibility and expressiveness. If the true governing equation includes terms that lie outside the predefined library, the method will fail to find the correct equation. Expanding the candidate library is a possible solution, but it cannot guarantee completeness and adds complexity to the sparse identification process. Therefore, one line of research focuses on developing methods to construct more flexible and adaptive libraries.

\subsubsection*{Expandable library methods} 
Expandable library methods expand the candidate library, allowing more expressive equations to be discovered. Evolutionary partial differential equation (EPDE)~\cite{maslyaev2019data} integrates evolutionary algorithms with sparse regression, enabling the discovery of governing equations by iteratively expanding the candidate library and dynamically exploring new terms. Compared to traditional sparse regression, EPDE can identify equation structures even under high levels of noise. Afterwards, DLGA-PDE~\cite{xu2020dlga} improves EPDE by integrating deep learning with genetic algorithms. In this framework, PDEs are digitized and encoded to form the corresponding genome. For example, the genome of the contaminant transport equation is
\begin{equation}\label{eq:dlga}
	u_t = -\mu u_x + D_Lu_{xx} \leftrightarrow [1], \{[1],[2]\}.
\end{equation}
Here, the orders of derivatives are represented by numbers, \emph{e.g.},  $u\leftrightarrow 0, u_x\leftrightarrow1, u_{xx}\leftrightarrow2$.  DLGA-PDE operates in two main steps. First, similar to other neural network-based methods, a DNN is learned to map the data and compute derivatives. Then, a genetic algorithm is applied to find the PDE. DLGA-PDE not only demonstrates efficacy on limited and noisy data with fast convergence but also shows flexibility in handling the left-hand side (time derivatives) of the PDE. Unlike most methods, which assume the left-hand side is always the first-order time derivative, DLGA-PDE can adapt to more general cases, \emph{e.g.}, the left-hand side is the second-order time derivative. A robust variant, R-DLGA, has been developed by incorporating PINNs~\cite{raissi2019physics} to improve accuracy under noisy and limited data~\cite{xu2021robust}. 

Beyond evolutionary approaches, PDE-Net 2.0~\cite{long2019pde} introduces a symbolic multi-layer neural network (SymNet), which leverages network topology to produce interaction terms, providing a hybrid numeric-symbolic framework for discovering time-dependent PDEs. Yu et al.\cite{xiangnan2025data} propose a fractional PDE discovery framework that incorporates fractional calculus into the library expansion process, enabling the identification of non-local memory effects in complex systems. 

Although evolutionary algorithms and frameworks such as PDE-Net are more flexible than fixed-library sparse regression in exploring candidate terms, they still do not span the full functional space. This limitation is not primarily due to the omission of particular operators, since additional functions can often be incorporated into the library. Rather, it reflects a deeper constraint of library-based representations: their basic units are still finite, predefined functional fragments. As a result, discovery is restricted to selecting and recombining a bounded set of building blocks, rather than constructing arbitrary equation structures from first principles.

This restriction becomes most apparent for expressions with richer compositional structure, such as composite functions, nested functions, or fractional forms. Unless such motifs are already encoded in the candidate library or genetic fragments, they cannot emerge naturally through recombination. The limitation is therefore intrinsic to the representational basis itself. Overcoming it requires a shift in the basic unit of representation, from fixed functional fragments to elementary operators and operands. This is precisely the transition made by symbolic approaches, which enable equations to be built compositionally and thereby enlarge the space of discoverable governing laws.

\subsubsection*{Open-form methods}
Open-form methods allow equations to be constructed symbolically, typically as hierarchical compositions of mathematical operators and variables. This substantially enlarges the space of possible governing equations, but also makes the search problem much harder: identifying the optimal symbolic structure from data is NP-hard\cite{virgolinsymbolic}. As a result, exhaustive search quickly becomes intractable as equation complexity grows, motivating the development of more efficient exploration strategies\cite{dong2025recent}. Examples include variable augmentation to compress complex subexpressions into auxiliary variables\cite{kahlmeyer2024scaling}, Monte Carlo tree search in Symbolic Physics Learner (SPL)\cite{sunsymbolic}, and large-scale parallel evaluation with subtree reuse in parallel symbolic enumeration (PSE)\cite{ruan2026discovering}.

Early applications of symbolic regression to differential equation discovery focused mainly on nonlinear dynamical systems and ODEs. Bongard and Lipson\cite{bongard2007automated} were among the first to use automated reverse engineering to infer governing equations directly from data, introducing strategies such as partitioning, automated probing, and snipping to improve scalability. Schmidt and Lipson later proposed a related framework based on matching numerically estimated derivative relationships from time-series data with their symbolic counterparts derived from candidate functions\cite{schmidt2009distilling}; this idea was subsequently implemented in the commercial software Eureqa\cite{dubvcakova2011eureqa}. Subsequent studies extended genetic programming to improve robustness and expressivity, including approaches based on Eulerian approximations\cite{gaucel2014learning}, epigenetic search mechanisms\cite{la2016inference}, and PDE-oriented variants such as SGA-PDE, which can identify equations with compound functions and fractional terms\cite{chen2022symbolic}. Related hybrid approaches have also combined symbolic search with gradient-based parameter estimation to improve PDE discovery under limited and noisy observations\cite{cohen2024physics}.

Progress in open-form discovery has also been shaped by major advances in symbolic regression more broadly. AI Feynman\cite{udrescu2020ai} shows that embedding physical priors and problem decompositions into symbolic search can dramatically improve the recovery of compact analytical expressions, whereas PySR\cite{cranmer2020discovering} demonstrates that large-scale, highly optimized symbolic regression can efficiently uncover interpretable equations with state-of-the-art accuracy. Although both methods were developed primarily for algebraic equation discovery rather than differential equations, they have had a substantial influence on the field by establishing scalable and physically informed paradigms for open-form equation search. At the same time, differential equation discovery poses additional challenges beyond algebraic regression, particularly the need to estimate derivatives from data and to represent differential operators explicitly. To address these issues, recent work has increasingly incorporated neural and learning-based components. For example, Lu \emph{et al.}\cite{lu2022discovering} combine a neural encoder with sparse symbolic modelling to discover ODEs and PDEs from partially observed systems. MechNNPDE~\cite{pervezmechanistic} leverages neural architectures to adaptively select and weight candidate terms, enabling the identification of governing equations beyond the limitations of fixed libraries.

Reinforcement learning provides a complementary route by framing equation generation as a sequential decision process. Deep symbolic regression (DSR)\cite{petersendeep} uses a recurrent neural network to generate candidate expressions and optimize them through reward-driven policy learning, and later extensions such as DISCOVER and R-DISCOVER adapted this strategy to ODE and PDE discovery, with improved robustness to scarce and noisy data\cite{du2024discover,du2024physics}. More broadly, the field has begun to shift from purely system-specific optimization towards pretrained and generative models that can transfer across families of equations. ODEFormer\cite{dodeformer}, for example, uses a transformer architecture trained on diverse equations to capture recurring structural patterns, enabling efficient adaptation to previously unseen systems. EqGPT similarly combines generative modelling with mathematical prior knowledge to construct and optimize free-form PDEs, and was reported to recover a previously unreported governing equation for strongly nonlinear surface gravity waves approaching breaking from experimental data\cite{xu2025generative}.

In parallel, efforts to learn embedding spaces for symbolic expressions, such as SNIP\cite{meidanisnip} and GenSR\cite{li2026gensr}, suggest a possible route towards more transferable and similarity-aware equation discovery, although these advances remain largely confined to algebraic settings. The emergence of large language models has extended this trend further: methods such as LLM4ED\cite{du2024llm4ed}, Scientific Generative Agent\cite{ma2024llm}, LLM-SR\cite{shojaeellm}, and LaSR\cite{grayeli2024symbolic} demonstrate how language-based priors and natural-language guidance can broaden the scope of symbolic discovery. Together, these developments point towards a new generation of open-form methods that are increasingly generative, transferable, and knowledge-guided.

\subsection*{Discovering differential equations with increasing coefficient complexity}
In addition to structural form, the coefficients of governing equations play a critical role in determining system behaviour. These coefficients may be broadly categorized into three classes: constant coefficients, coefficients expressible as deterministic functions of space or time, and coefficients that cannot be represented through explicit functional relations.

Constant coefficients are the simplest case and are well supported by most existing discovery frameworks. In many real-world systems, however, governing parameters exhibit spatial or temporal variability that reflects underlying heterogeneity or environmental influences. Such equation-expressible coefficients can, in principle, be inferred jointly with the governing operator, although doing so increases the dimensionality of the inference problem and introduces additional identifiability challenges.

More complex scenarios arise when coefficients correspond to stochastic or high-dimensional fields, as encountered in turbulent flows or porous media transport. These equation-inexpressible coefficients are often represented through random field expansions, such as the Karhunen–Loève expansion, and may not admit compact closed-form descriptions. In such cases, discovery methods must account for variability that cannot be captured by a single deterministic operator, further complicating the inference process.

\subsubsection*{Constant coefficients} 
Once the terms of the equation have been determined, the remaining task is to estimate the associated constant coefficients. This is usually a parameter estimation problem. When the equation is linear in the coefficients, these parameters can often be identified accurately using least-squares regression or related linear methods. When the dependence on the coefficients is nonlinear, more general iterative solvers, such as Newton-type methods, gradient-based optimization, or nonlinear least-squares procedures, may be required. Consequently, most of the existing literature, including library-based methods (\emph{e.g.}, PDE-FIND~\cite{rudy2017data}) and symbolic regression-based approaches (\emph{e.g.}, DISCOVER~\cite{du2024discover}), effectively address the discovery of constant coefficients. In fact, many differential equation discovery methods stop at this stage, as the identification and optimization of constant coefficients are computationally efficient and well-supported by current frameworks. However, this reliance on constant coefficients often limits the applicability of these methods to real-world problems, where coefficients frequently vary spatially, temporally, or stochastically, requiring more advanced approaches to address such complexities.

\subsubsection*{Equation-expressible coefficients} Many physical systems exhibit coefficients that vary systematically across space, time, or other variables. When such variation can itself be described through explicit functional relationships, we refer to these as \textit{equation-expressible coefficients}.
This setting is related to, but distinct from, the classical inverse problem in PDEs. In inverse problems, the governing equation is assumed to be known, and the goal is to infer unknown parameters or coefficient fields from observations. Differential equation discovery is more ambitious, aiming to identify both the active terms and the functional form of their associated coefficients.

A representative example is the framework of Rudy et al.~\cite{rudy2019data}, which extends sparse PDE discovery to the case of non-constant coefficients. Rather than fitting a single global coefficient vector, the method uses a grouped sparse regression strategy in which multiple data segments share a common equation structure while permitting the coefficient values to vary across time or space. This allows the method to capture systematic coefficient variation without abandoning the sparse representation of the governing law. Li et al.~\cite{li2020robust} further developed this idea by exploiting the low-rank structure often present in PDE-governed data, leading to a low-rank grouped sparse regression framework with improved robustness. Although these methods can effectively identify how coefficients change across conditions, they do not by themselves provide explicit closed-form expressions for the coefficient functions.

To recover explicit functional forms, Stepwise-DLGA~\cite{xu2021deep} adopts a winner-takes-all grouping strategy that partitions observations, tallies frequently selected terms across groups, and then infers symbolic expressions for the varying coefficients, reducing local overfitting at the cost of greater computational complexity. Other lines of work discretize the domain into fitting windows and treat coefficients as locally constant within each window, trading bias for variance control in noisy regimes. Beyond these, symbolic regression–based models such as stacked equation learners (SEQL) and hyper-EQL (HEQL) have been proposed to learn compact, interpretable formulas for space/time-varying coefficients while jointly identifying the governing operators~\cite{zhang2023deep}. 

GraphED\cite{xu2025beyond} introduces a graph-based representation, replacing the traditional tree-based structure, to enable the expression of equations with more complex coefficients. In this representation, operators and variables are encoded as nodes within a directed graph, while their relationships are defined by edges. Each edge carries a parameter that can represent either a constant value or a learnable coefficient, allowing the framework to capture richer parametric dependencies within the equation.

\subsubsection*{Equation-inexpressible coefficients} 
Unlike constant coefficients or equation-expressible coefficients, equation-inexpressible coefficients often correspond to stochastic or highly irregular fields, such as permeability in pollutant diffusion or thermal conductivity in heat transfer problems. These coefficients are commonly represented as random fields generated by techniques like Karhunen–Loeve expansion (KLE)~\cite{huang2001convergence,zhang2004efficient}, as shown in \Cref{fig:category}c. While such fields capture the variability and complexity inherent in real-world systems, they pose significant challenges for existing differential equation discovery methods. Traditional approaches, such as sparse regression or piecewise fitting, struggle with these coefficients due to their inability to account for strong nonlinearity or spatial irregularity. For example, fitting methods that assume constant coefficients within predefined windows often fail to generalize for large windows or overfit for narrow ones, particularly in scenarios involving intricate physical fields.

Closely related efforts have recently emerged in stochastic PDE discovery, where randomness enters through stochastic evolution or forcing terms rather than through irregular coefficient fields themselves~\cite{mathpati2024discovering}. However, equation discovery for PDEs with stochastic coefficient fields remains comparatively underexplored. One of the few methods developed specifically for this setting is the highly nonlinear parametric PDE discovery (HIN-PDE) framework~\cite{luo2023physics}, which employs a spatial kernel sparse regression model to estimate highly nonlinear coefficient fields. By introducing kernel-based coupling among neighboring spatial coefficients, HIN-PDE imposes local smoothness while retaining sufficient flexibility to capture strong spatial irregularity. This design improves robustness, mitigates overfitting, and enables more accurate identification of PDEs with complex coefficient fields from sparse, noisy, and irregularly sampled data.

Taken together, structural and coefficient complexity define the conditions under which governing equations can be discovered from finite observations. Methods that operate in regimes of low structural and coefficient complexity often succeed by imposing strong sparsity assumptions within predefined functional bases. As complexity increases along either dimension, however, successful discovery depends less on restrictive priors alone and more on the interplay among representation, optimization, and evaluation. The REO framework provides a unified perspective on this transition by clarifying how representational choices, search strategies, and evaluation criteria jointly determine the practical and fundamental limits of equation discovery across different physical regimes.

\subsection*{The ambiguous boundary between structural terms, coefficients, and noise}
A central but underappreciated issue in differential equation discovery is that the distinctions among structural terms, coefficients, and noise are often less fundamental than they appear. In current formulations, the structure of an equation is usually associated with its terms, whereas coefficients are treated as parameters and unexplained residual variation is assigned to noise. However, this partition is not naturally determined by the governing system itself. It is a consequence of how current algorithms formulate the discovery problem.

In principle, the goal of differential equation discovery is to infer the general governing equation in \Cref{eq:de}. In practice, however, directly searching such a broad functional space is infeasible for existing algorithms. As a result, many methods, especially sparse-regression and symbolic approaches, reformulate the task into the more restricted evolution form in \Cref{eq:de-evolution}. This reformulation decomposes discovery into the coupled problems of term selection and parameter estimation, thereby making the search computationally tractable. Such a decomposition is computationally useful, but it should not be mistaken for a fundamental property of the underlying system.

Indeed, a coefficient under one representation may become a discoverable structural relation once additional operating conditions, control variables, or regimes are introduced. Effective diffusivities, transport coefficients, and Reynolds-number-dependent closures are familiar examples of quantities that may appear as fixed parameters in a restricted setting, but reveal further law-like organization in a broader one. From this perspective, the boundary between structural complexity and coefficient complexity often reflects the representational limits of the algorithm as much as the structure of the system itself. The same point applies to noise. Residual mismatch is often treated as measurement error, but for an unknown system, it may also reflect missing terms, unresolved dynamics, imperfect coarse-graining, or a representation that is not yet aligned with the governing mechanism. More generally, robust discovery should not be viewed only as denoising before equation discovery, but as the problem of separating portable mechanism, condition-dependent parametrization, and irreducible variation. In this sense, part of what is labeled as noise may in fact correspond to structure that the current representation cannot yet recover.

The two-dimensional phase diagram introduced in this section provides an effective way to map the current algorithmic landscape of differential equation discovery. Compared with classifications based solely on method type, this view is more informative because it organizes algorithms according to the distinct tasks they are designed to handle. It offers a more task-oriented perspective on the field and makes the practical differences among methods easier to interpret. However, this picture remains tied to how existing algorithms formulate the problem.

In the next section, we move beyond this algorithm-dependent decomposition and introduce a representation-evaluation-optimization framework that emphasizes the common core of differential equation discovery. Rather than treating terms, coefficients, and noise as fixed ontological classes, this framework focuses on how candidate equations are represented, how they are evaluated, and how the resulting search is carried out.

\section*{The representation--evaluation--optimization framework: a review perspective}

The phase diagram above organizes differential equation discovery from the perspective of \emph{problem difficulty}, showing how discovery tasks vary with structural and coefficient complexity. A complementary question is how existing methods, despite their algorithmic diversity, can be understood within a common conceptual language. At a more abstract level, the objective of differential equation discovery can be understood through three fundamental questions. The first is a question of representation: how can the space of possible equations be expressed in a form that can be manipulated by computational methods? The second is a question of evaluation: by what criteria should a candidate equation be judged, given that success is rarely defined by predictive accuracy alone, but also by simplicity, physical plausibility, robustness, and solvability? The third is a question of optimization: how can one search efficiently through an immense and highly structured hypothesis space to identify equations that best satisfy these criteria? Together, these questions define the core intellectual structure of data-driven equation discovery.

\begin{figure}
	\centering
	\includegraphics[width=0.95\linewidth]{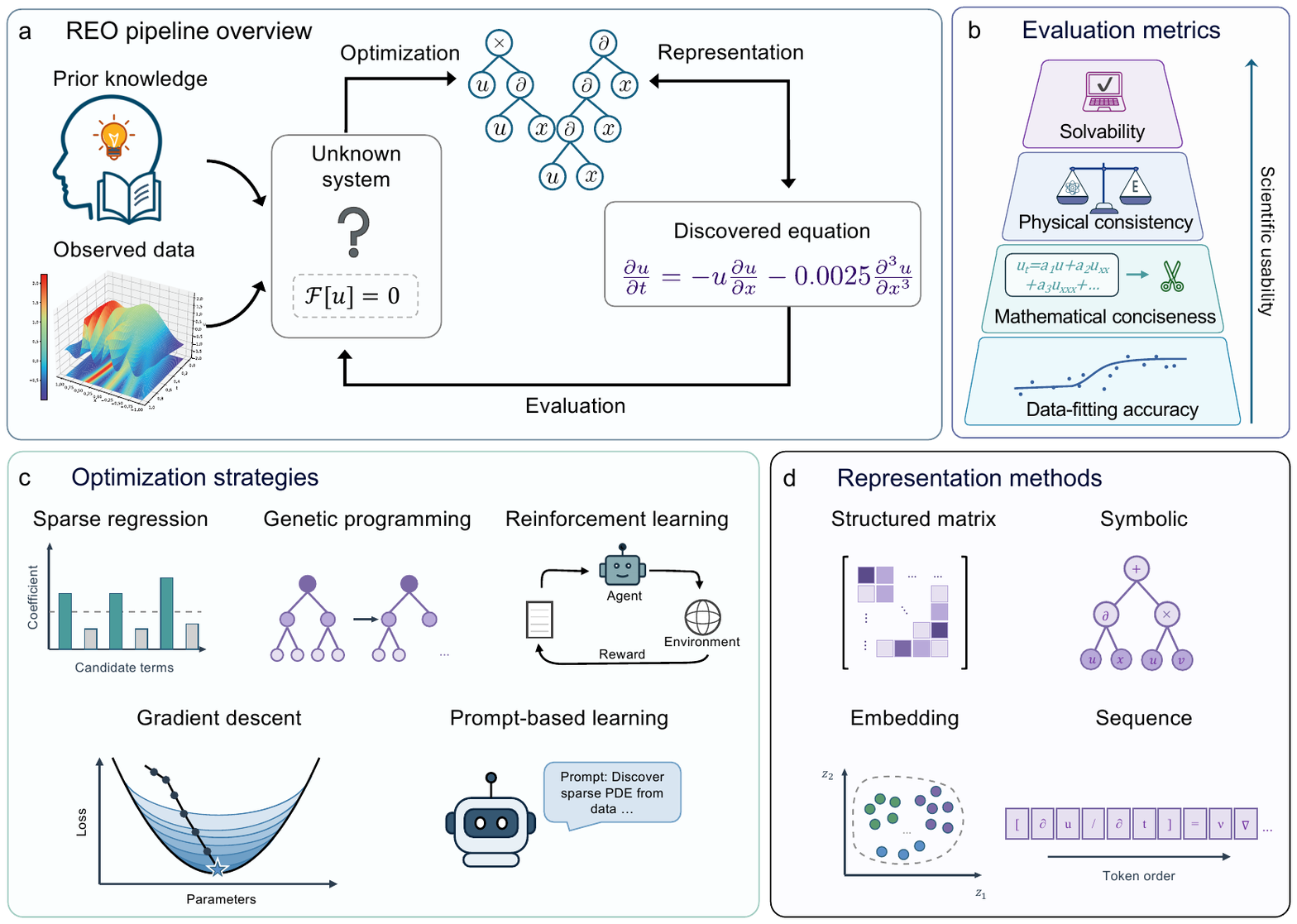}
	\caption{\textbf{Overview of the REO framework for discovering governing equations.} \textbf{a} Illustration of the REO pipeline, which integrates prior knowledge and observed data through representation, evaluation, and optimization to infer the governing equation. \textbf{b} The evaluation metrics for differential equation discovery, reflecting increasingly stringent criteria for scientific usability. \textbf{c} Representative optimization strategies commonly used in differential equation discovery, including sparse regression, genetic programming, reinforcement learning, gradient descent, and prompt-based optimization. \textbf{d} Representative equation representations, including structured matrix, symbolic, embedding, and natural language representations.}
	\label{fig:two_ascpects}
\end{figure}

In this review, we formalize this structure through the representation–evaluation–optimization (REO) framework, illustrated in \Cref{fig:two_ascpects}a. Rather than serving as a method-specific workflow, the REO framework provides an abstract decomposition of the equation discovery task into three interdependent components: The representation of candidate equations, the evaluation of candidate equations against data and scientific constraints, and the search strategies over candidate equations. This perspective provides a stable conceptual basis for navigating a rapidly evolving field, in which new methods continue to emerge across algorithms and applications.

\subsection*{Representation}
A key obstacle in equation discovery is that algorithms cannot interpret mathematical equations directly in the same flexible manner as human researchers. To make governing laws searchable by computational methods, candidate equations must first be converted into forms that can be represented, manipulated, and compared algorithmically. This defines the representation problem.

Formally, representation determines the hypothesis space over which governing equations are assumed to lie. In practice, symbolic equations are mapped into computationally tractable forms, such as structured matrices, traversal trees, high-dimensional embeddings, or string-based formulations, each of which implicitly restricts the class of admissible equations that can be inferred from data (\Cref{fig:two_ascpects}d).

\textbf{Structured matrix representation}, alternatively referred to as library-based representations, is widely used in differential equation discovery. Typically, a predefined set of candidate terms is constructed to represent the possible components of the governing equation. This set is often referred to as the library. For example, in SINDy~\cite{brunton2016discovering}, the library includes constants, polynomials (\emph{e.g.}, $x$, $x^2$), and trigonometric functions (\emph{e.g.}, $\sin(x)$). PDE-FIND~\cite{rudy2017data} extends the library to include derivatives of the state variables (\emph{e.g.}, $u_x$, $u_{xx}$) and their nonlinear interactions (\emph{e.g.}, $uu_x$), enabling the discovery of PDEs. This library is then represented as a structured matrix, where each row corresponds to a candidate term evaluated on the data, and each column corresponds to a specific data point, as shown in \Cref{fig:two_ascpects}d. 

Due to its conceptual simplicity and computational efficiency, this approach is widely used in real-world applications and is effective for systems with relatively simple dynamics. However, its performance is highly dependent on the completeness of the candidate library. If the target equation contains terms that are not included in the candidate library, the method will essentially fail to identify the correct equation. Efforts have been made to address this issue, such as incorporating physical priors, strategically extending the library to include more potential terms, or improving sparse regression algorithms to handle a large search space. However, the problem cannot be completely solved, as it is impossible to guarantee that the library contains all the true governing terms, especially in complex or unknown systems.

\textbf{Symbolic representations} provide an alternative approach to expressing differential equations. Specifically, one such representation is the traversal tree, where mathematical operations (\emph{e.g.}, addition, subtraction, multiplication, division, exponential, derivatives) are represented as internal nodes, while variables (\emph{e.g.}, $x, t, u$) and constants are represented as leaf nodes. In theory, any equation can be transformed into a symbolic form, such as a binary tree representation~\cite{chen2022symbolic}. The core idea of symbolic approaches is to leverage the structural relationships between symbols to express the operations that govern the equation. Symbolic representations are not limited to expression trees, but also include graph-based forms and genetic encodings. For example, methods such as DLGA~\cite{xu2020dlga} use gene-like encodings to represent mathematical relationships.

Symbolic representations are more flexible than structured matrix representations because they do not rely on a predefined library. Instead, they represent equations through finer-grained compositional units, rather than treating candidate terms as fixed objects. This more elementary representation enlarges the accessible hypothesis space by allowing a wider range of functional forms. Neural symbolic regression methods extend this idea by combining neural networks with symbolic encodings to efficiently infer governing equations from data, including networked or partially observed systems~\cite{ying2025neural}. However, this flexibility is achieved at the cost of greater search complexity. This trade-off reflects a general no-free-lunch principle in equation discovery: greater representational flexibility increases discovery potential, but inevitably makes optimization harder.

\textbf{Embedding representations} offer a fundamentally different approach by mapping equations into a continuous, high-dimensional vector space. Rather than preserving explicit symbolic structure, they prioritize forms that are efficient for computation and optimization. Although such representations are not directly interpretable by humans, they enable equations to be processed, compared, and manipulated efficiently by machine learning models, especially deep neural networks. Unlike the representations above, which are typically applied in a system-specific manner and often require retraining for each new equation system, embedding-based approaches can learn shared representations across systems. As a result, once trained, they can perform rapid inference on unseen systems and generalize across equation classes.

However, constructing a meaningful embedding space is nontrivial. For equation discovery, a latent space must satisfy two requirements\cite{li2026gensr}. First, it should be generative, so that continuously sampled latent vectors decode into syntactically valid equations. Second, it should support efficient search, such that local movements in vector space correspond, at least approximately, to meaningful changes in equation behaviour. These requirements are difficult to reconcile because symbolic structure and numerical behaviour do not align straightforwardly: expressions with distinct symbolic forms can be mathematically equivalent, such as $\sin^2 x$ and $1-\cos^2x$, whereas superficially similar terms can play very different roles in a dynamical model. For example, if the correct governing equation contains a fourth-order derivative, substituting a third-order derivative is not necessarily a better approximation than using a second-order derivative, because semantic similarity is determined by the function of an operator in the dynamics rather than by its algebraic order alone. As a result, local proximity in symbolic or algebraic space need not correspond to local proximity in behaviour.

\textbf{Sequence-based representations} have gained prominence in the era of large language models (LLMs)\cite{biggio2020seq2seq,shojaee2023transformer,vastl2024symformer,dodeformer,li2026gensr,du2024llm4ed}. Instead of constructing problem-specific embeddings from scratch, LLMs adopt a distinct paradigm in which equations are represented as serialized token sequences. Pretraining on large-scale text and code corpora endows these models with broad priors over symbolic syntax and compositional structure, making them a promising foundation for equation discovery.

Conceptually, LLM-based sequence representations are closely related to embedding representations, because the generated equations are still mediated by learned embeddings. What differs is the interface for discovery: embedding-based methods optimize directly in a continuous latent vector space, whereas LLM-based methods typically keep the model fixed and instead steer generation through prompts, context, and decoding. Observational data, prior knowledge and task instructions can therefore be provided directly in the prompt to guide equation generation.

The resulting expressions remain human-readable, which makes this representation well-suited to interactive and human-in-the-loop workflows. However, LLMs can also hallucinate, generating equations that appear plausible at the level of syntax or style but are mathematically incorrect, physically meaningless, or inconsistent with the observations. Sequence-based approaches, therefore, require careful validation and evaluation to ensure scientific reliability.

\subsection*{Optimization} 

In principle, if the chosen representation is expressive enough to encode the true governing equation, then equation discovery reduces to a search problem: with unlimited computational resources, the correct equation could eventually be identified by exploring the hypothesis space exhaustively. In practice, however, the set of admissible equations is effectively unbounded, making brute-force search infeasible. The central issue is therefore not only whether the correct equation is representable, but whether it can be found efficiently.

Optimization addresses this issue by governing how the search is carried out over the represented hypothesis space to identify governing equations that are consistent with observational data and prior physical constraints (\Cref{fig:two_ascpects}c). Accordingly, the effectiveness of an optimization strategy depends strongly on the structural properties of the chosen representation and the resulting search landscape.

\textbf{Sparse regression} aims to identify a parsimonious model that balances two competing objectives: structural simplicity and fidelity to the observed data. Specifically, it seeks a coefficient vector that is sparse while ensuring that the resulting right-hand-side expression in \Cref{eq:de-evolution} closely approximates the quantity on the left-hand side.

There is a rich literature on sparse-regression optimizers, including \(\ell_1\)-regularized formulations, such as Lasso and LARS \cite{tibshirani1996regression,efron2004least}, and elastic net \cite{zou2005regularization}, which balance sparsity and coefficient stability. Sequential threshold ridge regression (STRidge) \cite{rudy2017data} combines ridge regression with iterative thresholding to promote sparsity while maintaining numerical stability. Advanced methods, such as group and structured sparsity penalties \cite{yuan2006model,simon2013sparse,tibshirani2005sparsity}, extend the basic sparse regression framework by exploiting hierarchical and structural dependencies in the candidate terms. For small, well-defined libraries, exact \(\ell_0\)-regularization formulations using mixed-integer optimization have also been developed to ensure provably optimal solutions \cite{bertsimas2016best,hazimeh2020fast}. 

Sparse regression is particularly well suited to structured-matrix representations, where the hypothesis space is finite and explicitly parameterized by a fixed candidate library. In contrast, it is generally unsuitable for symbolic representations with unbounded or combinatorial search spaces, where the assumption of a finite linear dictionary no longer holds. Nevertheless, sparse regression is frequently adopted in the literature as a secondary optimization step for symbolic methods, where the equation structure has already been determined, and sparse regression is used to estimate or refine the associated coefficients. 

\textbf{Genetic algorithms} (GA) search for governing equations by evolving a population of candidate equations through selection, crossover, and mutation. Their action is especially transparent in tree-based representations, where equations are built from hierarchical combinations of operators, variables, and functions. Selection preserves better-performing candidates, crossover recombines subtrees from existing equations to form improved combinations of functional components, and mutation modifies local structures to introduce new possibilities. In this sense, crossover primarily refines combinations of existing building blocks, whereas mutation expands the search towards previously unexplored equation forms. Together, these operators enable GA to combine the exploitation of promising structures with the exploration of structural novelty.

GA is particularly effective when the search space is discrete, combinatorial, and non-smooth, as is often the case when the equation form itself must be discovered. Unlike coefficient-regression methods, which typically optimize weights or selections within a fixed candidate library, GA operates directly on the structure of the equation and can therefore explore both new combinations of known terms and new functional forms beyond a predefined library. Although most commonly associated with tree-based symbolic regression, the same evolutionary principle can be extended to string, graph, and other symbolic equation encodings, provided that suitable genetic operators are available. Previous literature on genetic algorithms includes foundational work by Holland~\cite{holland1992adaptation}, later systematic treatments by Goldberg~\cite{goldberg2013genetic}, and widely used introductory and tutorial expositions by Mitchell and Whitley~\cite{mitchell1998introduction,whitley1994genetic}.

In \textbf{reinforcement learning} (RL), the optimization process is formulated as a sequential decision-making problem, where an agent incrementally constructs a candidate equation by selecting operators, operands, or structural actions step by step. The agent is trained to maximize a reward signal that reflects how well the constructed equation fits the observed data, possibly augmented by penalties on complexity or invalid structures. The RL framework consists of states, actions, rewards, and a policy that governs the decision process.

RL is particularly suitable for structured generation problems with long-horizon dependencies and delayed rewards, where decisions made early in the construction process strongly influence the outcome. Similar to GA, RL is not inherently tied to a specific representation and can be applied to tree-based, graph-based, or string-based formulations, as long as the construction process can be expressed as a sequence of actions. However, the effectiveness of RL depends on the design of the state space, action space, and reward function, which must align with the chosen representation and task constraints. Representative reinforcement learning algorithms include value-based methods such as Q-learning~\cite{watkins1992q} and DQN~\cite{mnih2015human}, policy-based methods such as REINFORCE~\cite{williams1992simple}, and policy-optimization or actor--critic methods such as TRPO~\cite{schulman2015trust}, PPO~\cite{schulman2017proximal}, A3C~\cite{mnih2016asynchronous}, and SAC~\cite{haarnoja2018soft}.

\textbf{Gradient-based optimization} methods update model parameters by iteratively moving in the direction of the negative gradient of a differentiable loss function. Common variants include stochastic gradient descent (SGD)\cite{amari1993backpropagation}, momentum-based methods, and adaptive optimizers such as Adam~\cite{kingma2014adam} and RMSProp\cite{reddy2018optimization}. These algorithms are computationally efficient and scale well to high-dimensional continuous parameter spaces, making them the backbone of modern end-to-end learning systems.

Gradient descent is particularly well-suited to representations that admit differentiable parameterizations, such as embedding representations. When the mapping from parameters to model outputs is smooth and differentiable, gradient-based methods can provide efficient optimization and stable convergence. However, they are generally less suitable for discrete or combinatorial representations, such as symbolic expressions, where gradients are undefined or poorly behaved.

In the context of equation discovery, however, directly applying gradient descent raises several challenges. First, most gradient-based methods optimize primarily for predictive accuracy, which does not necessarily ensure that the learned representation captures the underlying physical or mathematical validity of the governing equation. As discussed above, symbolic structure and numerical behaviour do not align straightforwardly: equations with different symbolic forms can be mathematically equivalent, whereas superficially similar expressions can induce qualitatively different dynamics. As a result, minimizing data misfit alone may guide optimization toward embeddings that reproduce observations well but correspond to equations that are difficult to interpret, physically implausible, or structurally misleading. Second, the loss landscape in equation discovery is often highly non-convex, making gradient-based optimization sensitive to initialization and prone to poor local minima, especially when observations are noisy, sparse or partial. Third, gradient descent does not naturally enforce sparsity or symbolic interpretability, so accurate predictive models may still fail to decode into compact and scientifically meaningful equations. Recent efforts such as SNIP~\cite{meidanisnip}, EqGPT~\cite{xu2025generative}, and GenSR~\cite{li2026gensr} address these limitations through different modelling and optimization strategies, which are discussed later.

\textbf{Prompt-based optimization} has gained increasing attention with the emergence of LLMs, which shift part of the optimization problem from model parameters to prompt space. In this paradigm, performance is improved not by updating model weights, but by iteratively refining the prompt through in-context feedback, example-based conditioning, external evaluation signals, or model-generated critique. Representative strategies include automatic prompt generation and selection, optimization by prompting, prompt evolution, and textual-gradient-based refinement~\cite{zhou2022large,yang2023large,fernandopromptbreeder,yuksekgonul2024textgrad}.

Prompt-based optimization is often more accessible to human researchers because prior knowledge, physical constraints, and task-specific preferences can be injected directly through natural-language instructions\cite{yang2023large}. This makes it particularly appealing for interactive and human-in-the-loop scientific workflows\cite{shojaeellm}. However, prompt-based optimization also faces important limitations. First, it is an indirect form of control: rather than optimizing equations or model parameters directly, it steers the behaviour of a fixed pretrained model through textual instructions. The relationship between prompt changes and output quality is therefore often opaque and sensitive to small variations in wording. Second, prompt refinement does not by itself ensure mathematical validity, physical consistency, or agreement with the observed dynamics, so plausible-looking equations may still be incorrect or misleading. Third, effective prompt optimization usually depends on external evaluation signals, such as simulation error, symbolic checks, or model-based critique, which can make the overall procedure expensive and difficult to generalize. Prompt-based optimization, therefore, typically requires additional validation and feedback loops to support reliable equation discovery\cite{shojaeellm}.

\subsection*{Evaluation}

Given a candidate equation generated from the representation space, the evaluation stage determines its quality by assessing whether the resulting dynamics are consistent with observed system behaviour and whether the equation satisfies relevant physical constraints. Therefore, evaluation acts as the interface between \textit{representation} and \textit{optimization}, providing feedback signals that guide the optimization process toward physically plausible governing equations.

\subsubsection*{Quantitative evaluation metrics}
Evaluation metrics assess the quality of discovered equations from multiple perspectives. In this Review, as shown in \Cref{fig:two_ascpects}b, we group existing criteria into three main dimensions: \emph{data-fitting accuracy}, \emph{mathematical conciseness}, and \emph{physical consistency}. We further propose a fourth dimension, \emph{solvability}, which has been overlooked in the existing literature.

\textbf{Data-fitting accuracy} is a fundamental criterion for assessing how well a discovered equation explains the observed data. Two types of metrics are commonly used. First, regression-based metrics, such as the mean absolute error (MAE) or $R^2$ of the predicted target $u_t$, quantify how accurately the discovered differential equation reproduces the observed dynamics. These metrics reflect the overall predictive performance of the identified equation. Second, in benchmark settings where the ground-truth governing equation is known, term-level recall can be used to measure the proportion of true equation terms that are correctly identified. This metric evaluates the structural accuracy of the discovered differential equation.

\textbf{Mathematical conciseness} reflects the structural simplicity of the discovered equation. Simpler equations are often preferred because they are easier to interpret and align with Occam's razor. One way to measure conciseness is by counting the number of terms in the discovered equation, as fewer terms typically indicate a more streamlined representation of the system's dynamics. Another relevant metric is the symbolic length of the equation, which considers the total number of operators, variables, and coefficients. The Akaike information criterion (AIC) and Bayesian information criterion (BIC) are also standard measures that incorporate model complexity, providing a statistical basis for evaluating conciseness.

\textbf{Physical consistency} evaluates whether the discovered equations align with underlying physical principles. Xu et al.\cite{xu2023discovery} propose a physics-informed information criterion (PIC) to measure the parsimony and precision of the discovered PDE jointly. A key focus in the literature is improving discovery algorithms by embedding physical constraints. For instance, Ma et al.\cite{ma2024dimensional} highlight the importance of incorporating dimensional homogeneity to prevent the generation of physically invalid models. Similarly, PhysPDE~\cite{feng2025physpde} emphasizes bridging the gap between mathematical expressions and physical interpretability by incorporating physical hypotheses and theories (\emph{e.g.}, laws of physics and physical quantities) in PDE discovery.  AI-Hilbert\cite{cory2024evolving} inherently embeds constraints for physical consistency by expressing background knowledge as a system of polynomial equalities and inequalities. For instance, when deriving the Hagen-Poiseuille equation or the radiated gravitational wave power formula, the process strictly adheres to physical principles such as dimensional homogeneity. However, this area of evaluation remains underexplored, leaving room for further research to incorporate and systematically assess physical consistency in differential equation discovery frameworks. 

\textbf{Solvability} is proposed as a new dimension of evaluation, which has not yet been discussed in the literature. Solvability assesses whether the discovered equations can be effectively solved using analytical or numerical methods. Equations that are overly complex or stiff may fail to produce meaningful solutions, even if they fit the data well. By considering solvability, researchers can ensure that the discovered equations are practical for further analysis or simulation in real-world applications.

\subsubsection*{Benchmark datasets}
Beyond evaluation metrics, benchmark datasets, and canonical test systems provide controlled settings for assessing differential equation discovery methods. However, the extent to which performance on these benchmarks translates to real-world discovery remains an open question.

\textbf{Evaluation of ODE discovery methods.} 
Several datasets have been proposed to systematically assess the recoverability of governing equations by various ODE discovery methods. For example, Gennemark and Wedelin \cite{gennemark2009benchmarks} present a collection of more than $40$ benchmark problems for ODE model identification in cellular systems. The AI Feynman dataset collects 100 equations from the Feynman lectures on physics, offering a rich set of equations rooted in fundamental physical principles. However, this dataset mostly comprises algebraic equations, with only a few of them representing dynamical laws expressed as ODEs. Additionally, Gilpin\cite{gilpin2chaos} introduces the dysts standardized database of chaotic systems, which serves as a valuable resource for evaluating methods on highly nonlinear and chaotic dynamics. A class of SINDy algorithms is compared on this dataset, demonstrating its utility in benchmarking ODE discovery techniques\cite{kaptanoglu2023benchmarking}. The Strogatz dataset~\cite{la2021contemporary} contains 252 dynamic systems, focusing on one-dimensional systems. In contrast, ODEBench~\cite{dodeformer} proposes a more extensive dataset of 63 ODEs in dimensions one to four.

More recently, as LLMs have gained prominence in scientific discovery, the LLM-SRBench~\cite{shojaeellm} has been proposed. This comprehensive benchmark includes 239 challenging problems spanning four scientific domains (68 of which are ODE problems), and is specifically designed to evaluate LLM-based scientific equation discovery methods while minimizing the risk of trivial memorization. 

While these benchmarks provide useful environments for testing identifiability under controlled conditions, they typically rely on synthetic observations generated from known equations, and therefore do not fully reflect the challenges encountered in empirical discovery.

\begin{figure}
	\centering
	\includegraphics[width=\linewidth]{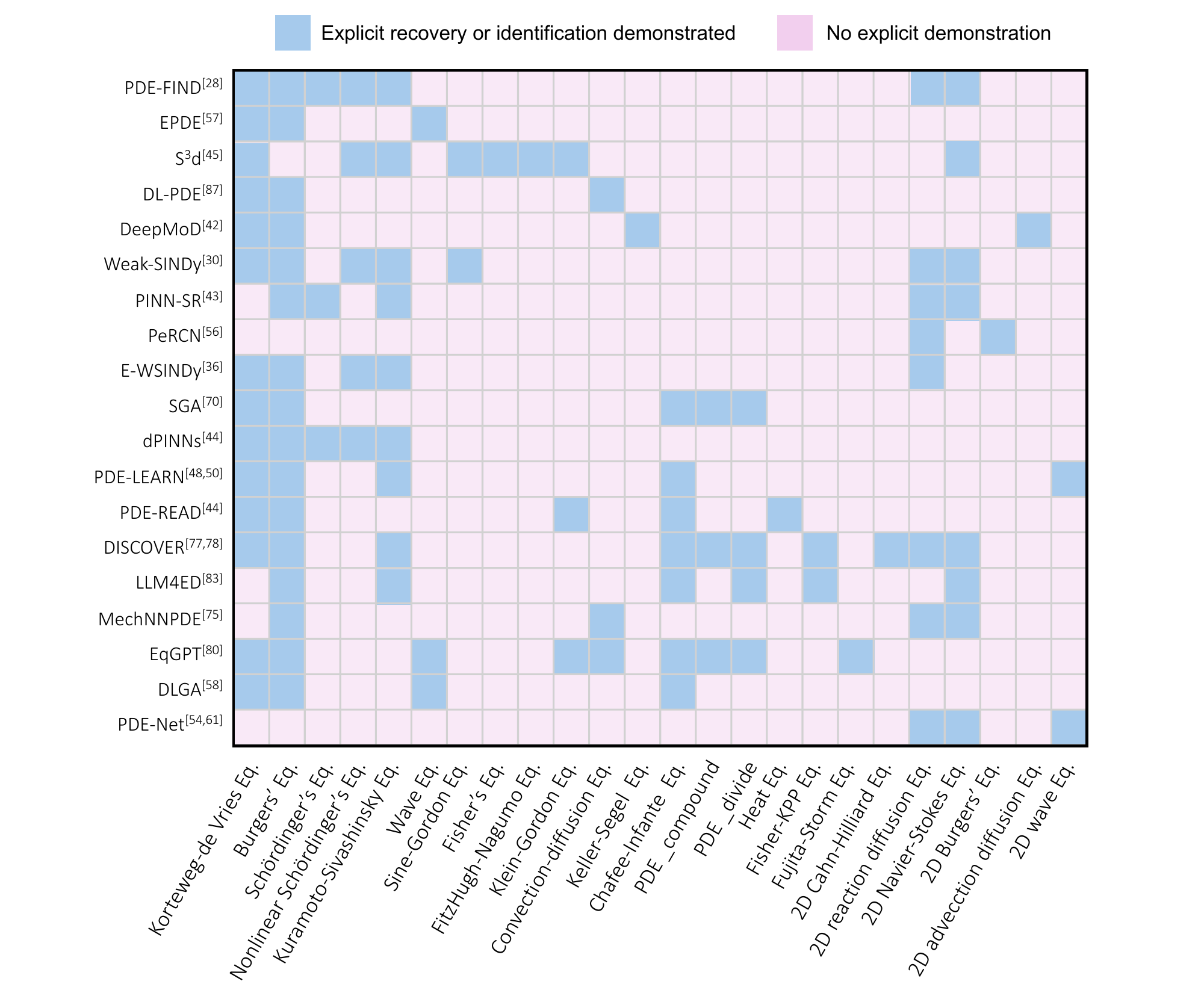}
	\caption{\textbf{Proof-of-concept validations reported across representative PDE discovery methods.} A blue box indicates that the paper explicitly demonstrates recovery or identification of the equation (or standard variants). A pink box indicates that the paper does not demonstrate this explicitly. The purpose of this table is to showcase the canonical PDEs used in various studies, rather than to evaluate the capability of the methods to recover these equations. Please refer to \Cref{tab:pde_list} in the Supplementary Information for the details of each PDE.}
	\label{fig:pde_methods}
\end{figure}

\textbf{Evaluation of PDE discovery methods.}
Rigorous evaluation of data-driven PDE discovery methods remains an open challenge. Despite rapid advancements in the field, current practices predominantly rely on bespoke \textbf{proof-of-concept} demonstrations, rather than standardized and comparable benchmarks. A common approach in the evaluation process involves testing methods on a limited set of well-known canonical PDEs, such as the Burgers' equation and Kuramoto–Sivashinsky equation (please refer to \Cref{tab:pde_list} for some commonly used canonical PDEs). These equations are typically simulated under controlled discretizations and carefully chosen boundary or initial conditions. While such benchmarks provide useful insights into the capabilities of discovery methods, they suffer several limitations: (1) It is difficult to directly compare the performance of different methods, since there are no standardized datasets, and researchers often generate their own synthetic data, which can vary significantly due to differences in boundary conditions, initial conditions, or numerical schemes. (2) The narrow focus on canonical PDEs limits the generalizability of these methods to more complex, domain-specific systems. 

Here, we provide a table (see \Cref{fig:pde_methods}) that summarizes which canonical PDEs have been used to validate different methods. Our intention is not to make direct comparisons but rather to illustrate the diversity of equations employed across studies. This overview aims to provide insights into the current evaluation practices and identify potential gaps for future research.

Although a few recent efforts have begun to develop dedicated benchmarks for PDE discovery, such as MDBench~\cite{bideh2026mdbench}, such resources remain limited. At present, there is still no widely adopted benchmark specifically designed for PDE discovery. As a result, several general-purpose PDE benchmarks, including PDEArena~\cite{guptatowards}, PDEBench~\cite{takamoto2022pdebench}, DynaDojo~\cite{bhamidipaty2023dynadojo}, and RealPDEBench~\cite{hu2026realpdebench} are often used, at least in part, for this purpose.

\subsubsection*{Toward rigorous and standardized evaluation}
To summarize, significant effort is still required to improve the evaluation of differential equation discovery methods. This calls not only for comprehensive and standardized benchmark datasets that include real-world systems from diverse domains, but also for standardized evaluation protocols that ensure meaningful comparison across methods. At present, many approaches are assessed on a case-by-case basis, which limits the field’s ability to judge generalizability. In particular, evaluation protocols should specify how noise and data scarcity are introduced. For noise robustness, this includes both the \emph{distribution} of the perturbation, such as Gaussian or uniform noise, and the \emph{mode} of injection, such as additive perturbations, \emph{i.e.}, \(\mathrm{observation}+\mathrm{noise}\), or multiplicative perturbations, \emph{i.e.}, \(\mathrm{observation}(1+\mathrm{noise})\). For sparse-data settings, commonly adopted sampling schemes, such as random sampling or Latin hypercube sampling, should likewise be standardized. Without such shared protocols, differences in reported performance may be driven as much by experimental design as by algorithmic capability.

A non-public test set is also important to reduce the risk of data leakage, especially for methods involving LLMs. In parallel, the field must prioritize reproducibility through open access to datasets, code, and evaluation pipelines. This will allow the community to identify which classes of differential equations are already tractable for current methods and which remain challenging. Finally, evaluation must move beyond fitting accuracy alone. Metrics assessing conciseness/interpretability, physical consistency, solvability, and computational cost are all needed to evaluate whether a method is not only accurate, but also scientifically useful and practically deployable. A systematic evaluation framework that combines benchmark datasets, transparent protocols, and diverse metrics will provide a much stronger basis for measuring progress and guiding future research.

\subsection*{Practical value of the REO framework}

The REO framework provides a conceptual lens for understanding how differential equation discovery is evolving. Specifically, the field has advanced through \emph{representational refinement}, moving from closed-form equation representations to more flexible open-form representations. As the basic building blocks of equation discovery become increasingly elementary, for example, shifting from predefined terms to operators, the space of possible equations expands correspondingly. A useful analogy is in physics: matter can be decomposed from molecules to atoms to quarks, revealing progressively more fundamental constituents. Similarly, decomposing equations into smaller, more primitive units increases their expressive power, allowing a broader range of dynamical laws to be represented. However, greater expressivity does not automatically translate into easier discovery. Richer equation spaces produce many candidate equations that fit the same data equally well, making the challenge twofold: searching efficiently through a vastly larger space, and identifying the solutions that are mechanistically meaningful.

This issue remains insufficiently understood in the current literature, where many methods are still developed from the perspective of improving representation, optimization, or evaluation in isolation. However, effective discovery depends on the interplay among all these three parts. Although empirical studies have explored optimization strategies tailored to different representations, a more rigorous mathematical foundation for this interplay is still lacking. For example, there is little research on the geometry and complexity of the latent equation spaces induced by different representations, the criteria for separating competing equations within those spaces, and the optimization principles required to search such spaces in a theoretically grounded manner.

\section*{From canonical to real-world systems}\label{sec:5}

The evaluation practices discussed above are based largely on synthetic datasets generated from canonical dynamical systems for which the governing equations are known \emph{a priori}. Although such benchmarks provide controlled settings for testing structural recoverability, they mainly assess whether a method can rediscover established laws under idealized conditions. The more demanding scientific objective is to infer governing relations in empirical systems, where measurements are noisy, variables may be only partially observed, and the underlying dynamics remain only partly understood.

In this section, we review applications of data-driven differential equation discovery through four recurring paradigms: learning from noisy and incomplete real-world data, bridging microscopic rules and macroscopic laws, discovering interpretable closure models, and exploring new governing relations. The aim is not to provide a comprehensive survey, as in-depth domain-specific reviews are available elsewhere\cite{sanderse2025scientific,song2024towards}. Instead, we draw representative examples from different disciplines to illustrate how these paradigms recur across fields, and to highlight the common challenges and opportunities that arise when equation discovery moves from idealized benchmarks to real scientific data.

\subsection*{Discovering equations from noisy and incomplete data}

A central scientific problem in data-driven equation discovery is how to infer reliable governing equations when observations are sparse, noisy, or incomplete. Such conditions are the norm in empirical science: forcing terms may be unmeasured, state variables may be hidden, experimental control may be limited, and noise may reflect either measurement uncertainty or intrinsic stochasticity. The value of equation discovery in this regime is therefore not simply algorithmic robustness, but the ability to extract interpretable physical laws from the imperfect data through which real systems are observed.

In \textit{experimental physics}, a key challenge is to recover fluid dynamical laws when important physical quantities cannot be directly measured. Weakly turbulent thin-layer flows provide such a setting, because forcing and pressure fields are often unavailable even though they strongly influence the observed dynamics. Reinbold et al.\cite{reinbold2020using,reinbold2021robust} address this problem by reconstructing governing equations from incomplete flow measurements, showing that interpretable dynamical laws can still be identified when essential variables are missing. This demonstrates that equation discovery can be used not only to fit measured fields, but also to infer the effective structure of unresolved physical processes.

A closely related problem arises when the observable data do not contain the full state of the system. In many physical and biological systems, only partial and noisy measurements are available, while the variables that close the dynamics remain hidden. Course et al.\cite{course2023state} address this setting by jointly estimating latent states and unknown governing equations, thereby linking state reconstruction with model discovery rather than assuming full observability. Lu et al.\cite{lu2022discovering} develop a similar strategy for partially observed systems, combining state reconstruction with sparse symbolic modelling to recover interpretable dynamics directly from incomplete observations. Similarly, Gao et al.\cite{gao2022autonomous} propose an autonomous framework to infer complex network dynamics from incomplete and noisy measurements, demonstrating that even under significant data sparsity and partial observability, the underlying network structure and governing equations can be accurately recovered. These studies show that equation discovery can be extended from fully observed benchmark systems to realistic settings in which the measured variables are only projections of a higher-dimensional dynamical process.

In \textit{electromagnetics}, the scientific difficulty is to identify governing relations when the measured fields are affected by unknown noise sources, parasitic couplings, and cross-physics interactions that are hard to specify from first principles. Xiong et al.\cite{xiong2019data} address this problem by combining deep learning with sparse regression to uncover hidden relations and coupling terms directly from field data. The scientific value is that equation discovery can expose effective electromagnetic interactions that are not explicitly encoded in a prior model, providing an interpretable route to characterize complex devices and media under measurement uncertainty.

This capability is especially important in \textit{geosciences}, where limited observational access and the impossibility of controlled experiments often prevent the direct application of classical model-identification approaches. In heterogeneous soil-water systems, the problem is to infer flow equations from field observations rather than idealized laboratory measurements. Song et al.\cite{song2025deep} address this challenge with the Extended-DeepGS framework, which identifies governing equations for soil-water flow directly from observational data. Similarly, groundwater systems pose the problem of recovering subsurface flow laws from sparse field measurements while preserving basic physical constraints. The PHY-PDE framework\cite{zhan2024physics} addresses this by integrating mass conservation with SINDy to infer groundwater equations from field data. These examples illustrate how equation discovery can translate irregular environmental observations into physically interpretable models of subsurface transport.

Finally, noisy data are not always a nuisance: in many systems, randomness is part of the underlying dynamics. The scientific problem is then to distinguish measurement noise from intrinsic stochastic forcing and to recover governing laws that explicitly encode random effects. Gao et al.\cite{gao2024learning} address this problem by inferring hidden stochastic differential equations from experimental data. This extends interpretable equation discovery beyond deterministic dynamics, showing that symbolic laws can still be recovered when stochasticity is an essential component of the system rather than an external perturbation.

\subsection*{Bridging microscopic rules and macroscopic behaviours}
In many complex systems, the available data come from particles, molecules or individual cells, whereas the scientific questions concern collective fields, transport laws or population-level dynamics. Data-driven equation discovery can therefore act as a scale-bridging tool: it identifies continuum or coarse-grained governing equations directly from high-dimensional microscopic data, without requiring the macroscopic form to be specified in advance.

This paradigm is particularly clear in \textit{active matter}, where collective motion emerges from local interactions among self-driven units. The scientific challenge is to determine which continuum equations describe these emergent flows when the microscopic rules are known only through simulations or experiments. Supekar et al.\cite{supekar2023learning} address this problem by learning hydrodynamic equations directly from particle simulations and experimental data, thereby translating agent-level interaction rules into continuum PDEs. The value of this result is not only predictive: it provides an interpretable macroscopic description of how local activity, alignment and interactions generate collective active flows.

A related scale-bridging problem arises in \textit{granular matter}, where macroscopic flow behaviour emerges from contact, friction and collision at the grain scale. Because these interactions are discrete and history-dependent, deriving continuum equations from first principles remains difficult. Zhao et al.\cite{zhao2023data} address this by discovering macroscopic governing equations for granular flows from discrete element method data. This links particle-resolved simulations to continuum-level transport laws, providing a route to interpretable models of dense granular motion without prescribing the constitutive structure in advance. The SiMBA framework\cite{de2025simplest} illustrates a similar principle for complex kinetic systems, where high-dimensional microscopic or mesoscopic behaviour is distilled into compact ODE models that retain the dominant coarse-grained dynamics.

In \textit{chemistry}, the scale-bridging problem is to connect fine-scale molecular or reaction dynamics to reduced descriptions that remain accurate over longer time and length scales. Arbabi et al.\cite{arbabi2020linking} address this by learning multiscale closures directly from fine-scale simulations, thereby constructing explicit links between models at different levels of description. In \textit{plasma physics}, the analogous challenge is to extract reduced models from fully kinetic simulations, where the microscopic phase-space dynamics are too costly to use directly in large-scale prediction. Data-driven discovery has been used to identify such reduced plasma models while balancing interpretability, accuracy and computational efficiency\cite{alves2022data}. These examples show how equation discovery can compress fine-scale dynamics into tractable governing laws without discarding the physical mechanisms that control the large-scale behaviour.

This scale-bridging role is especially valuable in \textit{biology}, where mechanisms span molecular, cellular and tissue scales. A central scientific problem is to explain how stochastic microscopic events, such as molecular interactions or single-cell gene-expression fluctuations, give rise to reproducible population-level or tissue-level dynamics. Gene expression programming has been used to infer macroscopic equations from molecular simulations\cite{xing2022using}, providing coarse-grained laws that summarize the collective consequences of molecular-scale rules. In single-cell biology, the challenge is to infer dynamical laws of cell-state transitions from static, noisy sequencing snapshots. The bi-level BILLIE framework\cite{li2025bi} addresses this by discovering RNA and protein velocity equations, thereby linking intracellular gene-expression kinetics to population-level differentiation dynamics. More broadly, such approaches address a central problem in biology: how stochastic single-cell behaviour can generate robust continuum-scale phenomena\cite{ducos2025evaluating}.

\subsection*{Discovering interpretable closure models}

Another major application area is closure modelling: the governing framework is often known, but key constitutive, subgrid, or unresolved terms cannot be derived or measured directly. This situation arises when the known equations describe the large-scale dynamics, while the effect of unresolved scales must be represented through additional relations. In this regime, equation discovery is valuable not because it replaces established physical laws, but because it can identify explicit and interpretable expressions for the missing terms, bridging empirical closure design and black-box machine learning.

In \textit{fluid dynamics}, the central closure problem is to represent the influence of unresolved turbulent fluctuations on the resolved mean flow. In the Reynolds-averaged Navier--Stokes equations, this difficulty appears through the Reynolds-stress tensor, whose exact form depends on turbulent correlations that are not directly available from the mean fields. Classical closures encode these effects through empirical or semi-empirical assumptions, whereas many machine-learning closures improve accuracy without yielding transparent physical structure. Data-driven equation discovery addresses this problem by searching for explicit algebraic closure relations from turbulence data. Beetham and Capecelatro\cite{beetham2020formulating} identify interpretable RANS closures for homogeneous free-shear turbulence and internal channel flows, and Beetham et al.\cite{beetham2021sparse} extend this strategy to multiphase flows. Similarly, the SpaRTA framework\cite{schmelzer2020discovery} discovers algebraic Reynolds-stress models that improve RANS predictions while preserving equation-level interpretability. These studies show that closure discovery can convert high-fidelity turbulence data into usable constitutive relations for reduced-order flow models.

Related closure problems arise when the missing terms are not only turbulent stresses but also effective forcing, boundary effects, or transformations between flow regimes. In wind-energy applications, the scientific challenge is to model turbine wakes without resolving all blade-scale and wake-scale interactions. The SRDWM framework\cite{wang2025symbolic} addresses this by discovering equation-level closures for volumetric forcing and boundary terms, yielding interpretable wake models that can be embedded in larger-scale flow simulations. In compressible turbulence, the problem is to relate velocity and temperature profiles across regimes with different thermodynamic structure. Zhang et al.\cite{zhang2026transformations} address this by discovering symbolic transformations that map compressible turbulent profiles onto incompressible forms. These examples illustrate how closure discovery can expose compact relations between resolved variables and unresolved physical effects, rather than merely improving predictive fits.

In \textit{geophysics}, closure problems are especially tied to multiscale dynamics, because climate and Earth-system models evolve large-scale fields whose tendencies depend on unresolved mesoscale or microscale processes\cite{sanderse2025scientific}. In ocean dynamics, a central problem is to represent the momentum transport induced by mesoscale eddies without explicitly resolving them. Zanna and Bolton\cite{zanna2020data} address this by discovering an interpretable parameterization of subgrid momentum fluxes, thereby linking unresolved eddy effects to resolved large-scale flow variables. In atmospheric modelling, cloud cover presents a related problem: it is controlled by unresolved thermodynamic and microphysical processes, yet it strongly affects radiation and circulation at larger scales. Grundner et al.\cite{grundner2024data} address this by combining symbolic regression with physical constraints in a hierarchical framework to discover diagnostic equations for cloud cover from model data. In both cases, the scientific value lies in completing existing geophysical models with explicit closures that encode unresolved processes in a form that can be inspected, tested and reused.

\subsection*{Toward discovering novel governing laws}

The most ambitious scientific goal of data-driven equation discovery is to identify governing structures that are not already contained in established models. In this regime, the problem is not merely to estimate parameters, choose closures or improve numerical predictions, but to expose mathematical relations that may point to new mechanisms. This possibility is especially important in fields where experiments and simulations generate large datasets, yet mechanistic theory remains incomplete or difficult to derive from first principles.

In \textit{chemistry and biochemistry}, a central challenge is to infer reaction mechanisms, empirical laws and structure--property relations from complex experimental data. Traditional discovery often depends on trial-and-error exploration, which can be slow when the space of molecular structures, synthesis conditions or material compositions is large. Data-driven equation discovery addresses this problem by searching directly for explicit relations that connect measurable inputs to chemical or material outcomes\cite{ingelsten2025data}. Such approaches can accelerate materials discovery by identifying interpretable structure--property relationships\cite{ghosh2023discovery,cole2020design,cole2021shape,lou2026unsupervised}, and can also support the construction of more accurate and efficient computational models for chemical systems\cite{demir2023recent}. A concrete example is provided by Xu et al.\cite{xu2025explicit}, who discover an explicit coupling relation between thin-layer chromatography and column chromatography from experimental measurements. The scientific value is that a practical experimental correspondence, usually treated empirically, is converted into an explicit mathematical relation that can be inspected and reused.

In \textit{geophysics}, the difficulty is often to identify effective macroscopic laws for systems whose full physics involve unresolved heterogeneity, complex boundary conditions or multiscale forcing. Viscous gravity currents provide such a problem: they are central to geophysical transport, yet simplified theory-derived equations may fail to capture the full effective dynamics observed in data. Zeng et al.\cite{zeng2023deep} address this by discovering macroscopic PDEs for viscous gravity currents, with the reported equations outperforming conventional theory-derived models in both short- and long-time prediction regimes. A related problem appears in precipitation--terrain interactions, where rainfall patterns reflect coupled atmospheric, topographic and hydrological processes that are difficult to reduce to a closed-form law. Xu et al.\cite{xu2025exploring} report explicit equations linking precipitation patterns to terrain characteristics, illustrating how equation discovery can reveal environmental relations that are hard to obtain directly from first principles.

In \textit{complex systems modelling}, the scientific problem is broader still: the governing variables, operators and nonlinear terms may all be uncertain. Oceanic rogue waves provide one example, because rare extreme events arise from nonlinear wave interactions that are difficult to encode in compact predictive laws. Häfner et al.\cite{hafner2023machine} address this by using machine-guided symbolic discovery to derive a real-world model for rogue waves directly from data. EqGPT extends this idea to strongly nonlinear surface gravity waves approaching breaking, identifying a previously unreported PDE structure\cite{xu2025generative}. In both cases, equation discovery is used not simply to reproduce wave data, but to propose interpretable mathematical forms for regimes where conventional modelling remains incomplete.

Similar opportunities arise in engineered and biological transport systems. In traffic dynamics, the scientific challenge is to infer the hidden continuum laws and source terms that govern congestion, propagation and dissipation on networks. TRAFFIC-PDE-LEARN\cite{wei2025neural} addresses this problem by discovering PDE structures directly from network traffic data, moving beyond the calibration of preselected traffic-flow models. In biological transport, the problem is to determine which taxis mechanisms actually drive observed bacterial migration. Psarellis et al.\cite{psarellis2024data} use PDE discovery to infer active taxis mechanisms in \textit{E.~coli} migration, thereby moving beyond standard Keller--Segel formulations toward experimentally grounded mechanistic interpretation. Finally, Gu et al.\cite{gu2025discover} extend the scope of symbolic discovery from governing equations to analytical solution objects by uncovering explicit Green's functions for differential operators. This shows that data-driven discovery can also reveal interpretable mathematical objects that support analysis, not only simulation or prediction.

\section*{Discussion and outlook}\label{sec:6}

\begin{figure}
	\centering
	\includegraphics[clip, trim=0cm 7cm 0cm 0cm,width=\linewidth]{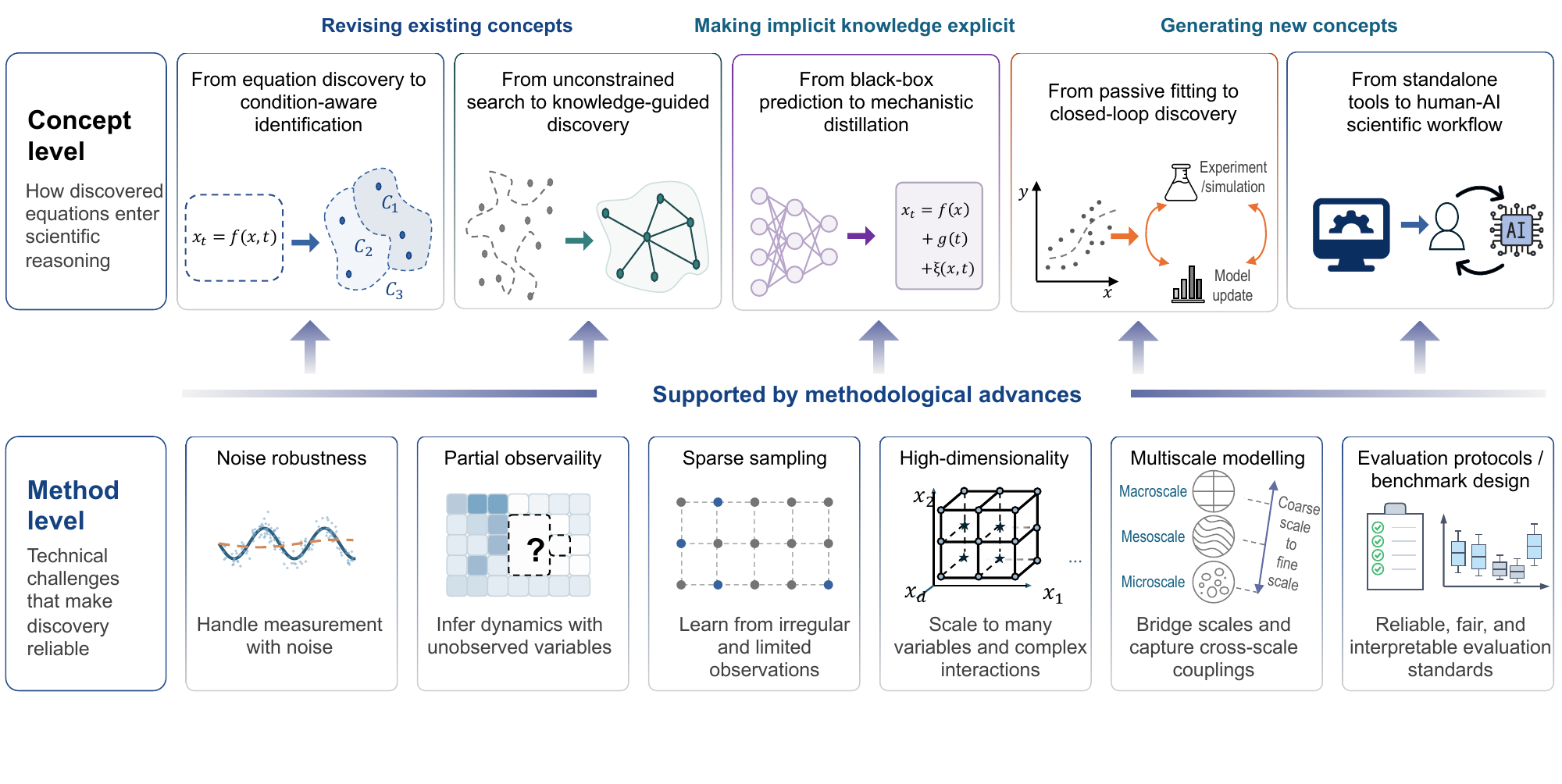}
	\caption{\textbf{Outlook for differential equation discovery: from recovering equations to forming scientific concepts.}
The future development of differential equation discovery may be understood at two levels. At the methodological level, important challenges remain in dealing with noisy, partially observed, sparsely sampled, high-dimensional, and multiscale systems, as well as in establishing reliable evaluation protocols. At a broader level, the field may shift from equation recovery towards condition-aware system identification, knowledge-guided discovery, closed-loop discovery with adaptive data generation, mechanistic distillation of black-box models, and human–AI scientific workflows.}
	\label{fig:future}
\end{figure}

Differential equation discovery is emerging as a useful tool for extracting governing equations from complex systems. However, an equation is rarely important in science only because it fits the observed data. It becomes scientifically meaningful when it changes how a system is understood. It may reveal a missing mechanism, formalize an intuition, or define a new object for theoretical analysis. For example, the Laplacian first appears as a differential operator, but over time it becomes a concept for describing how local differences drive spatial change. Therefore, equation discovery should move beyond symbolic expression generation toward the identification of structures that can become part of scientific reasoning.

From this perspective, the field of differential equation discovery still remains as an initial step. Most existing work is concerned with recovering individual equations under controlled settings, often with synthetic or benchmark data, and much of the methodological progress has focused on algorithmic improvements such as more flexible representations or more efficient optimization strategies. These advances are important, but they do not address the more fundamental and difficult question of how a discovered equation becomes scientifically meaningful. Here we do not attempt to list all technical directions, as outlined in \Cref{fig:future}. Instead, we organize the outlook around the question: how can differential equation discovery contribute to the formation of scientific concepts?

\noindent\textbf{Revising existing concepts under real conditions.}
Many scientific applications do not require rediscovering an entire governing equation from scratch. The main structure may already be known. What is unknown is how that structure should be completed in a particular regime, such as the relevant boundary conditions, closure terms, control variables, or unresolved couplings. Equation discovery is then a route to improve an existing concept rather than replace it.

This is important because established concepts are often idealizations. They are derived under assumptions that may fail in real-world systems, such as complex materials, turbulent flows, heterogeneous media, and multiscale systems. Therefore, a discovered correction term or closure relation can be more than a technical improvement to prediction. It may indicate how the original concept should be improved. For example, a missing transport term may point to an overlooked mechanism, and a learned closure may suggest a new effective interaction. The discovered term helps researchers rethink the limitations of the original model and suggest directions for improvement.

This also changes the role of prior knowledge. Rather than treating equation discovery as an unconstrained symbolic search, future methods should use existing theory as a scaffold. Dimensional constraints, symmetries, conservation laws, admissible operator classes and known constitutive forms should guide the search from the beginning. In this setting, prior knowledge defines the conceptual space in which a discovered equation can be interpreted. Pretrained generative models may eventually extend this idea by learning reusable priors over equation structures, as suggested by early work such as EqGPT. The crucial issue, however, is whether such models can generate physically meaningful equations, not only statistically plausible.

\noindent\textbf{Turning scientific intuition into mathematical form.}
A second route to concept formation concerns knowledge that scientists already use but cannot easily formalize. Researchers often rely on physical intuition, modelling experience, and qualitative ideas about mechanisms when they choose variables or simplify a system. This knowledge is important, but it is not always written down as an equation. Equation discovery may help make part of it explicit.

This is especially relevant for scientific machine learning. Neural networks can make accurate predictions, but what they have learned is usually hard to express in a form that scientists can directly reason about. Equation discovery offers one way to open this black box. This perspective, which may be viewed as \emph{mechanistic distillation}, is valuable for several reasons. First, it can expose what a black-box model has learned in a form that is more accessible to physical reasoning. Second, it can compress high-parameter predictive systems into lower-dimensional equation-based surrogates, potentially reducing computational cost and enabling deployment in resource-constrained settings. Third, it can bridge the gap between predictive accuracy and mechanistic interpretability, thereby supporting the broader adoption of such models in scientific and engineering practice. In this sense, beyond being a tool for recovering laws from nature, equation discovery can also be regarded as a way to white-box scientific machine-learning models.

\noindent\textbf{Producing new concepts through closed-loop discovery.}
The most ambitious possibility is that equation discovery may help generate concepts that are not already present. This will not happen from a single passive fit. A new concept usually emerges when the same mathematical structure appears across multiple situations. Discovery, therefore, has to become more iterative.

This motivates a shift from passive fitting to closed-loop discovery. In many applications, the limiting factor is not the total volume of data. A single experiment or simulation can produce large spatiotemporal fields. The limitation is the number of independent regimes. Under these circumstances, many candidate equations may remain equally plausible. The next observation should therefore be chosen not only to reduce prediction error, but to distinguish between competing equations. Active learning, adaptive experiments, and high-throughput simulations can play this role by turning equation discovery into a cycle: propose candidate structures, identify where they disagree, collect targeted data, and revise the model.

More importantly, equation discovery serves as a critical intermediate layer in automated research. Modern scientific practice increasingly generates massive datasets through high-throughput experiments and automated simulation platforms. Within this vast data, new concepts can emerge, but only if the patterns are translated into a form suitable for reasoning. Equation discovery converts this raw output into symbolic representations that can be implemented in computational frameworks, enabling human scientists to perform mechanistic reasoning, derivation, and ultimately concept formation. In this way, human–AI collaboration becomes a loop: AI proposes candidate equations, humans interpret and validate them, and together they uncover structures that drive scientific understanding forward.

Together, these perspectives highlight that differential equation discovery is a central tool for structuring physical knowledge. Connecting raw experimental and simulation data with symbolic representations, it enables both refinement of existing concepts and the creation of new ones. In doing so, equation discovery bridges high-dimensional measurements, mechanistic reasoning, and formal theoretical analysis, functioning as a key enabler of human–AI collaboration and the next generation of physics research.


\bibliography{sample}

\section*{Acknowledgements}
This work was financially supported by the National Natural Science Foundation of China (Nos. 12572266, 12501744, 92270118) and Zhejiang Provincial Natural Science Foundation of China (No. LD25F020001). Besides, this work was also supported by the High-Performance Computing Centers at Eastern Institute of Technology, Ningbo, and Ningbo Institute of Digital Twin.

\section*{Author contributions}

Yuntian Chen conceptualized the review topic and designed the overall structure of the manuscript. Siyu Lou conducted the literature search, analyzed and organized the references, and drafted the initial sections of the manuscript. Hao Xu contributed to the methodological overview and helped synthesize key developments in equation discovery methods. Wenguan Wang and Linfeng Zhang contributed to the discussion of applications and future research directions. Lu Lu contributed to the development of the conceptual framework and helped refine the organization of the review. Yang Liu and Dongxiao Zhang contributed to content refinement, critical revision, and proofreading of the final manuscript. All authors participated in discussions, reviewed the manuscript, and approved the final submission.

\section*{Competing interests}
The authors declare no competing interests. 

\section*{Key points}

\noindent\textbullet~Data-driven differential equation discovery is moving from canonical equation recovery to real-world scientific inference in noisy, partial, and multiscale systems.

\noindent\textbullet~Discoverability depends jointly on equation-structure complexity and coefficient complexity, which determine what data, priors, and optimization strategies are needed.

\noindent\textbullet~Representation, evaluation, and optimization provide a common language for comparing methods beyond algorithmic labels.

\noindent\textbullet~Real applications increasingly use equation discovery to infer hidden dynamics, bridge microscopic and macroscopic descriptions, and construct interpretable closure models.

\noindent\textbullet~The next challenge is to turn recovered equations into scientific concepts that revise theories, distill mechanisms, and guide closed-loop experiments.

\clearpage
\section*{Supplementary information}

\renewcommand{\thetable}{S\arabic{table}}
\setcounter{table}{0}

\section*{List of ODE benchmark systems}
For convenience and ease of reference, we provide a detailed list of the systems included in commonly used ODE benchmark datasets. \Cref{tab:ode_benchmark_overview} provides a high-level summary of commonly used ODE benchmark datasets, including their size, typical dimensionality, primary focus, and main limitations. We also include direct links to the associated repositories for ease of navigation.

\begin{longtable}{p{2.7cm} p{1.0cm} p{1.5cm} p{2.0cm} p{3.0cm} p{4.0cm}}
\caption{High-level summary of representative benchmark datasets for ODE discovery. The table highlights differences in scope, typical dimensions and benchmarking purpose across commonly used datasets.}
\label{tab:ode_benchmark_overview}\\
\toprule
\textbf{Benchmark} & \textbf{Year} & \textbf{Size} & \textbf{Typical dimension} & \textbf{Main focus} & \textbf{Remarks} \\
\midrule
\endfirsthead

\toprule
\textbf{Benchmark} & \textbf{Year} & \textbf{Size} & \textbf{Typical dimension} & \textbf{Main focus} & \textbf{Remarks} \\
\midrule
\endhead

\midrule
\multicolumn{6}{r}{Continued on next page}
\endfoot

\bottomrule
\endlastfoot

Gennemark--Wedelin benchmark suite$^{*,}$\cite{gennemark2009benchmarks}
& 2009
& $>40$
& low to moderate
& ODE model identification in cellular systems
& An early biologically motivated benchmark collection, particularly useful for system-identification tasks in systems biology. \\

\href{https://sites.google.com/site/biopredynbenchmarks/}{BioPreDyn-bench}\cite{villaverde2015biopredyn}
& 2015
& 6
& moderate to high
& Parameter estimation and kinetic modelling in systems biology
& A systems-biology benchmark suite focused on challenging parameter-estimation problems rather than broad ODE discovery.\\

\href{https://space.mit.edu/home/tegmark/aifeynman.html}{AI Feynman}\cite{udrescu2020ai}
& 2020
& 100
& mostly low-dimensional expressions
& Symbolic regression on physics-inspired formulas
& Useful as a benchmark for scientific symbolic regression, but only a small subset of the expressions correspond to dynamical laws. \\

\href{https://github.com/williamgilpin/dysts}{\texttt{dysts}}\cite{gilpin2chaos}
& 2021
& 131
& mostly 2--4
& Chaotic and highly nonlinear dynamical systems
& particularly valuable for testing robustness on chaotic ODEs, and widely used in recent benchmarking studies. \\

\href{https://github.com/lacava/ode-strogatz}{Strogatz dataset}\cite{la2021contemporary}
& 2021
& 252
& mostly 1--2
& Canonical textbook dynamical systems
& Broad coverage of interpretable low-dimensional systems, with especially strong representation of one-dimensional dynamics. \\

\href{https://github.com/pdebench/PDEBench}{ODEBench}\cite{dodeformer}
& 2024
& 63
& 1--4
& Standardized benchmark for ODE discovery
& Designed as a modern benchmark with broader dynamical coverage than many earlier ODE collections. \\

\href{https://github.com/deep-symbolic-mathematics/LLM-SR}{LLM-SRBench}\cite{shojaeellm}
& 2025
& 239 total \newline (68 ODE tasks)
& mixed
& Benchmarking LLM-based scientific equation discovery
& Includes a dedicated ODE subset and is explicitly designed to reduce success through trivial memorization. \\

\href{https://github.com/CP3-Origins/cp3-bench}{cp3-bench}\cite{thing2025cp3}
& 2025
& 7
& mixed
& Symbolic regression benchmarking in cosmology and astroparticle physics
& A benchmarking tool for comparing symbolic regression algorithms on cosmology-inspired datasets; relevant as a neighboring SR benchmark, but not a dedicated ODE-discovery benchmark. \\

\href{https://github.com/gryaklab/mdbench}{MDBench}\cite{bideh2026mdbench}
& 2026
& 63 ODEs \newline (+ 14 PDEs)
& mixed
& Data-driven ODE/PDE discovery
& A recent benchmark framework for model discovery, with substantial ODE coverage and systematic evaluation under noise; broader than ODE-only collections because it also includes PDE tasks. \\

\multicolumn{6}{p{\linewidth}}{\footnotesize $^{*}$ The original benchmark website, \texttt{www.odeidentification.org}, appears to be outdated.}
\end{longtable}

\section*{List of PDE benchmark systems}
For convenience and ease of reference, we provide a list of canonical PDE systems that are frequently used in benchmarking studies for PDE discovery. These equations span several major classes, including transport and diffusion equations, dispersive and wave equations, reaction--diffusion systems, higher-order pattern-forming PDEs, and fluid-dynamical models. They cover both scalar and coupled systems in one and two spatial dimensions, and therefore reflect many of the core challenges in PDE discovery, such as nonlinear advection, higher-order derivatives, multi-field coupling, and complex spatiotemporal behaviour. \Cref{tab:pde_list} summarizes these representative systems and their standard forms.

\begin{longtable}{p{5cm}p{9cm}}
\caption{Representative canonical PDEs commonly used in PDE discovery benchmarks.}
\label{tab:pde_list}\\
\toprule
\textbf{PDE} & \textbf{Form} \\
\midrule
\endfirsthead

\toprule
\textbf{PDE} & \textbf{Form} \\
\midrule
\endhead

\midrule
\multicolumn{2}{r}{Continued on next page} \\
\endfoot

\bottomrule
\endlastfoot

Korteweg--de Vries equation & $u_t + 6u u_x + u_{xxx} = 0$ \\

Burgers' equation & $u_t + u u_x - 0.1u_{xx} = 0$ \\

Schr\"odinger equation & $i u_t + \frac{1}{2}u_{xx} - \frac{x^2}{2}u = 0$ \\

Nonlinear Schr\"odinger equation & $i u_t + \frac{1}{2}u_{xx} + |u|^2u = 0$ \\

Kuramoto--Sivashinsky equation & $u_t + u u_x + u_{xx} + u_{xxxx} = 0$ \\

Wave equation & $u_{tt} - \frac{1}{4}u_{xx} = 0$ \\

Sine--Gordon equation & $u_{tt} - u_{xx} + \sin(u) = 0$ \\

Fisher's equation & $u_t - u + u^2 - 0.1u_{xx} = 0$ \\

FitzHugh--Nagumo equation &
$\left\{
\begin{aligned}
u_t - u_{xx} - u(u - 0.2)(1 - u) - w &= 0, \\
w_t - 0.002u + 0.001w &= 0
\end{aligned}
\right.$ \\

Klein--Gordon equation & $u_{tt} - u_{xx} + u + u^3 = 0$ \\

Convection--diffusion equation & $u_t + u_x - 0.25u_{xx} = 0$ \\

Keller--Segel equation &
$\left\{
\begin{aligned}
u_t - 0.5u_{xx} + 10u w_{xx} + 10u_x w_x &= 0, \\
w_t - 0.5w_{xx} + 0.05w - 0.1u &= 0
\end{aligned}
\right.$ \\

Chafee--Infante equation & $u_t - u_{xx} + u - u^3 = 0$ \\

PDE Compound & $u_t - (u u_x)_x = 0$ \\

PDE Divide & $u_t + \frac{1}{x}u_x - 0.25u_{xx} = 0$ \\

Heat equation & $u_t + 0.05u_{xx} = 0$ \\

Fisher--KPP equation & $u_t - 0.02u_{xx} - 10u + 10u^2 = 0$ \\

Fujita--Storm equation & $u_t - 0.05\left(\frac{u_x}{u^2}\right)_x = 0$ \\

2D Cahn--Hilliard equation & $u_t - (u^3)_{xx} - (u^3)_{yy} + u_{xx} + u_{yy} + 0.5(u_{xx})_{xx} + 0.5(u_{yy})_{yy} = 0$ \\

2D reaction--diffusion equation &
$\left\{
\begin{aligned}
u_t - 0.1u_{xx} - 0.1u_{yy} + uv^2 + u^3 - v^3 - u^2v - u &= 0, \\
v_t - 0.1v_{xx} - 0.1v_{yy} + uv^2 + u^3 + v^3 + u^2v - v &= 0
\end{aligned}
\right.$ \\

2D Navier--Stokes equation & $\omega_t + u\omega_x + v\omega_y - 0.01\omega_{xx} - 0.01\omega_{yy} = 0$ \\

2D Burgers' equation &
$\left\{
\begin{aligned}
u_t + u u_x + v u_y - 0.005(u_{xx} + u_{yy}) &= 0, \\
v_t + u v_x + v v_y - 0.005(v_{xx} + v_{yy}) &= 0
\end{aligned}
\right.$ \\

2D advection--diffusion equation & $u_t - 0.5u_{xx} - 0.5u_{yy} + 0.25u_x + 0.25u_y = 0$ \\

2D wave equation & $u_{tt} - u_{xx} - u_{yy} = 0$ \\

\end{longtable}

\subsection*{Open-source softwares}
In addition to benchmark datasets, reproducible software implementations play an important role in the evaluation of differential-equation discovery methods. We therefore summarize representative open-source packages and official repositories that are commonly used for ODE and PDE discovery in \Cref{tab:de_codebases}.

\begin{longtable}{p{3.1cm} p{2.4cm} p{1.2cm} p{1.4cm} p{2.9cm} p{4.0cm}}
\caption{Representative publicly available repositories for differential-equation discovery.}
\label{tab:de_codebases}\\
\toprule
\textbf{Tool / repository} & \textbf{Method family} & \textbf{Scope} & \textbf{Language} & \textbf{Main use} & \textbf{Remarks} \\
\midrule
\endfirsthead

\toprule
\textbf{Tool / repository} & \textbf{Method family} & \textbf{Scope} & \textbf{Language} & \textbf{Main use} & \textbf{Remarks} \\
\midrule
\endhead

\midrule
\multicolumn{6}{r}{Continued on next page} \\
\endfoot

\bottomrule
\endlastfoot

\multicolumn{6}{l}{\textit{General-purpose ODE/PDE discovery packages}} \\
\href{https://github.com/dynamicslab/pysindy}{PySINDy}
& sparse regression / SINDy
& ODE/PDE
& Python
& Sparse discovery of governing equations
& Widely used package centered on SINDy and related methods. \\

\href{https://github.com/SciML/DataDrivenDiffEq.jl}{DataDrivenDiffEq.jl}
& structural identification 
& ODE/PDE
& Julia
& Data-driven model discovery
& Part of the SciML ecosystem for structural estimation and identification of differential equations. \\

\makecell[l]{\href{https://github.com/PhIMaL/DeePyMoD}{DeePyMoD} \\\href{https://github.com/PhIMaL/DeePyMoD_torch}{DeePyMoD\_torch}}
& neural model discovery
& ODE/PDE
& Python
& Discovery from noisy data
& Modular framework combining neural surrogates, feature libraries, and sparse selection for ODE/PDE discovery. \\

\href{https://github.com/menggedu/EDL}{EDL}
& LLM-assisted equation discovery
& ODE/PDE
& Python
& Automatic discovery of nonlinear dynamics
& Research code for a large-language-model-based framework for automatic equation discovery of nonlinear dynamical systems. \\

\multicolumn{6}{l}{\textit{ODE-focused repositories}} \\
\href{https://github.com/MilesCranmer/pysr}{PySR}
& symbolic regression
& ODE-related
& Python + Julia
& Symbolic equation discovery
& General symbolic-regression package often used as a baseline or component for ODE discovery, though not ODE-specific. \\

\href{https://github.com/sdascoli/odeformer}{ODEformer}
& transformer-based symbolic regression
& ODE
& Python
& Symbolic regression of dynamical systems
& A transformer-based method for discovering ODE systems from data. \\

\href{https://github.com/dynamicslab/SINDy-PI}{SINDy-PI}
& implicit sparse regression
& ODE
& MATLAB
& Discovery of implicit/rational dynamics
& Parallel implicit SINDy, useful for rational nonlinearities and implicit dynamics. \\

\href{https://github.com/andreikitaitsev/SymINDy}{SymINDy}
& symbolic + sparse regression
& ODE
& Python
& Reconstruction of strongly nonlinear systems
& Extends SINDy with combinatorial search over elementary functions. \\

\href{https://github.com/anvari94/IRK-SINDy}{IRK-SINDy}
& sparse regression + implicit RK
& ODE
& Python
& Discovery from sparse/corrupted data
& Open-source implementation combining SINDy-style regression with implicit Runge--Kutta ideas and neural components. \\

\href{https://github.com/Rose-STL-Lab/symmetry-ode-discovery}{Symmetry-informed ODE discovery}
& symmetry-aware discovery
& ODE
& Python
& Equation discovery with symmetry priors
& Official codebase for symmetry-informed governing-equation discovery, with baselines including SINDy, weak SINDy, and PySR-based GP. \\

\href{https://github.com/albertotonda/symbolic-regression-ode-systems}{symbolic-regression-ode-systems}
& symbolic regression
& ODE
& Python
& ODE system identification
& Research code focused on learning ODE systems via symbolic regression and trajectory-to-regression transformations. \\

\href{https://github.com/peterparity/symder}{SymDer}
& neural-symbolic discovery
& ODE
& Python
& Discovery from partial observations
& Combines latent-state reconstruction with sparse symbolic dynamics. \\

\multicolumn{6}{l}{\textit{PDE-focused repositories}} \\
\href{https://github.com/snagcliffs/PDE-FIND}{PDE-FIND}
& sparse regression / PDE-FIND
& PDE
& Python
& Classical PDE discovery
& Canonical repository for PDE-FIND; the repo itself notes that more modern implementations are available in PySINDy. \\

\href{https://github.com/MathBioCU/WSINDy_PDE}{WSINDy\_PDE}
& weak-form sparse regression
& PDE
& MATLAB
& PDE discovery in weak form
& Official-style repository for weak SINDy for PDEs. \\

\href{https://github.com/dm973/WSINDy_PDE_OL}{WSINDy\_PDE\_OL}
& online weak-form sparse regression
& PDE
& MATLAB
& Online PDE discovery
& Repository for online weak-form sparse identification of PDEs. \\

\href{https://github.com/ITMO-NSS-team/EPDE}{EPDE}
& evolutionary equation discovery
& PDE
& Python
& Flexible symbolic PDE discovery
& Dedicated framework for equation discovery beyond fixed linear libraries. \\

\href{https://github.com/menggedu/DISCOVER}{DISCOVER}
& reinforcement learning / symbolic search
& PDE
& Python
& Open-form PDE discovery
& Reinforcement-learning framework for discovering concise PDEs with limited prior structural assumptions. \\

\href{https://github.com/YuntianChen/SGA-PDE}{SGA-PDE}
& evolutionary symbolic regression
& PDE
& Python
& Open-form PDE discovery
& Symbolic genetic algorithm for discovering PDEs without fixed equation templates. \\

\href{https://github.com/m2lines/EquationDisco}{EquationDisco}
& hybrid linear/symbolic regression
& PDE
& Python
& Discovery over spatial fields
& Focused on equation discovery over 2D periodic spatial fields with symbolic operators. \\

\href{https://github.com/woshixuhao/EqGPT}{EqGPT}
& generative equation discovery
& PDE
& Python
& Symbolic PDE discovery
& Public code accompanying a generative framework for PDE discovery that learns reusable mathematical patterns from handbook-style equations and supports free-form term generation. \\

\end{longtable}

\subsection*{Discretization}
To validate the differential equation discovery method as a proof-of-concept demonstration, one often begins with a known DE. Consider a system in $n$-dimensional space $\mathbf{x}\in\mathbb{R}^n$ and time $t\in\mathbb{R}$. Each spatial dimension is discretized into $k_i$ points, and time is discretized into $m$ points. The complete spatialtemporal grid consists of $N=\prod_{i=1}^nk_i$ spatial points and $m$ temporal points. The state variable $u(\mathbf{x},t)$ is discretized into a $(n+1)$-dimensional tensor $\mathcal{U}\in \mathbb{C}^{k_1\times k_2\times \dots \times k_n \times m}$, where the first $n$ dimensions correspond to the spatial discretization and the last dimension corresponds to time. For implementation purposes, the tensor is often reshaped into a matrix format. Let $\mathbf{U}\in \mathbb{C}^{K\times m}, K=\prod_{i=1}^kn_i$ be the flattened representation where each row corresponds to a spatial location and each column to a time instance:
\begin{equation}
\mathbf{U} = \begin{bmatrix}
u(\mathbf{x}_1, t_1) & u(\mathbf{x}_1, t_2) & \cdots & u(\mathbf{x}_1, t_m) \\
u(\mathbf{x}_2, t_1) & u(\mathbf{x}_2, t_2) & \cdots & u(\mathbf{x}_2, t_m) \\
\vdots & \vdots & \ddots & \vdots \\
u(\mathbf{x}_N, t_1) & u(\mathbf{x}_N, t_2) & \cdots & u(\mathbf{x}_N, t_m)
\end{bmatrix},
\end{equation}
where $\{\mathbf{x}_j\}_{j=1}^K$ enumerates all spatial grid points.

\begin{tcolorbox}[mybox, title=Box 1: PDE-FIND~\cite{rudy2017data} overview, label=box:method]
PDE-FIND considers PDEs of the following form:
\begin{equation}\label{eq:pdefind}
\mathbf{u}_t = \sum_{k} \theta_k\, \Phi_k\big(\mathbf{u},\nabla \mathbf{u},\nabla^2 \mathbf{u},\ldots\big),
\end{equation}
The key assumption in PDE-FIND is that $\{\theta_k\}$ is \textit{sparse}, meaning that only a small subset of terms in the library $\Phi$ contributes significantly to the dynamics.

\textbullet~\textbf{Representation.} Upon discretization, \Cref{eq:pdefind} can be expressed in the following form:
\begin{equation}\label{eq:library}
\mathbf{U}_t = \Theta(\mathbf{U}, \mathbf{Q})\theta,
\end{equation}
where $\theta = [\theta_1, \theta_2, \dots, \theta_k]$ are the \textit{constant parameters} to each term in the library $\Theta$, which includes $k$ terms.  The goal is to identify the sparse vector $\theta$. 

\textbullet~\textbf{Evaluation} PDE-FIND was tested on six benchmark PDEs in a proof-of-concept manner (\Cref{fig:pde_methods}). The numerical solution was obtained using spectral differentiation and a Runge-Kutta-45 ODE solver. Specifically, to simulate real-world applications, artificial noise (white noise with a magnitude equal to 1\% of the solution function's standard deviation) was added to the solution. To compute derivatives from the data, PDE-FIND used \textit{polynomial interpolation}. At each grid point, a local polynomial of degree \( P \) was fit to the neighboring data points, and derivatives of the polynomial were used to approximate the derivatives of the data. This method was particularly effective in mitigating the impact of noise while maintaining accuracy. Nevertheless, boundary points remain challenging due to the lack of neighboring data, which can contribute to errors in derivative computation.

\textbullet~\textbf{Optimization.} Optimizing \Cref{eq:library} is challenging due to the high dimensionality of the candidate library $\Theta$. To this end, PDE-FIND introduced STRidge, which alternated between regression and thresholding to enforce sparsity. First, ridge regression was used to minimize the residual error while penalizing large coefficients:
\begin{equation}
\theta = \arg \min_{\hat{\theta}} \| \Theta(\mathbf{U,Q}) \hat{\theta} - \mathbf{U}_t \|_2^2 + \lambda \| \hat{\theta} \|_2^2,
\end{equation}
where \( \lambda \) is the ridge regularization parameter. Next, coefficients smaller than a predefined threshold \( \varepsilon \) were set to zero:
\begin{equation}
\theta_i = 0 \quad \rm{if} \quad |\theta_i| < \varepsilon.
\end{equation}
The remaining active terms were then refit using ridge regression, and the process was repeated iteratively until convergence.

\end{tcolorbox}

\begin{tcolorbox}[mybox, title=Box 2: Reinforcement-based differential equation discovery (DSR~\cite{petersendeep}/DISCOVER~\cite{du2024discover}), label=discover]
    DSR was originally designed for algebraic equations; however, it can be adapted to discover the right-hand-side expression for ordinary differential equations (ODEs) of the form $u_t = f(u)$. That said, it cannot handle equations involving derivatives on the right-hand side. In contrast, DISCOVER extends this capability to differential equations of the general form discussed in \Cref{eq:pdefind}.
	
	\textbullet~\textbf{Representation.}  
	Symbolic expressions are represented as sequences derived from the pre-order traversal of their expression trees. In DSR, internal nodes correspond to mathematical operators (\emph{e.g.}, \(+, -, \times, \div\)), while leaf nodes represent operands (\emph{e.g.}, \(u, x, t\)). DISCOVER extends this framework by incorporating differential operators, such as $\partial, \partial^2$ as additional internal nodes, enabling the discovery of differential equations. 

    \textbullet~\textbf{Evaluation}  
	DSR benchmarks symbolic regression tasks on the Nguyen test suite, relying on synthetic datasets with both clean and noisy data. These datasets are used to evaluate candidate mathematical expressions for their fitness to the data. Artificial noise is added to test robustness, with Gaussian noise proportional to the root-mean-square of the dependent variable.  
	
	DISCOVER extends this by verifying its accuracy and efficiency on five benchmark PDEs as shown in \Cref{fig:pde_methods}. Artificial noise is similarly introduced, with $10\%$ noise for the KdV, Burgers', and Chafee-Infante equations, and $1\%$ noise for PDE\_divide and PDE\_compound. 
	
	DISCOVER addresses the challenge of numerical differentiation by employing automatic differentiation through a fully connected feedforward neural network, which maps system inputs (\emph{e.g.}, spatial coordinates $x$ and time $t$) to the state variable \(u\). This approach helps smooth noisy data but may risk overfitting when data quality is poor. To mitigate this, R-DISCOVER incorporates PINNs, embedding discovered equations as constraints into the neural network. 
    
	\textbullet~\textbf{Optimization.} The optimization process employs an agent, such as a recurrent neural network (RNN) or long short-term memory (LSTM) network, to generate equations. This agent is trained using a risk-seeking policy gradient, which prioritizes high-reward expressions. The policy gradient is computed as:
	\[
	\nabla_\theta J_{\rm{risk}} (\theta) = \mathbb{E}_{\tau \sim p(\tau | \theta)} \left[ R(\tau) \cdot \nabla_\theta \log p(\tau | \theta) \right],
	\]
	where \(\tau\) is a generated symbolic expression and \(R(\tau)\) is its reward. 
    
    In the DSR framework, the reward is defined as
    \begin{equation*}
        R = \frac{1 }{1 + \rm{NRMSE}}, \quad \rm{NRMSE} = \frac{1}{\sigma}\sqrt{\frac{1}{N}||U_t - \hat{U}_t||^2},
    \end{equation*}
    where $N$ is the number of samples, $\sigma$ is the standard deviation of the left-hand-side values $U_t$, $\hat{U}_t$ represents the predicted values from the right-hand-side equation. 
    
    DISCOVER extends this reward function to account for the hierarchical structure of PDEs, balancing data fitness and parsimony:
	\[
	R = \frac{1 - \zeta_1 k - \zeta_2 d}{1 + \rm{RMSE}}, \rm{RMSE} = \sqrt{\frac{1}{N}||U_t - \hat{U}_t||^2}
	\]
    where $k$ is the number of function terms of the governing equation, $d$ is the depth of the generated PDE expression tree, and $\zeta_1$ and $\zeta_2$ are penalty factors that encourage simplicity.

\end{tcolorbox}

\begin{tcolorbox}[mybox, title=Box 3: ODEFormer overview\cite{dodeformer}, label=discover]
	ODEFormer considers ODEs of the following general form:
	\begin{equation}\label{eq:odeformer}
		u_t = f(u), \quad \text{for some }\mathcal{F}:\mathbb{R}^n \mapsto \mathbb{R}^n.
	\end{equation}

	\textbullet~\textbf{Representation.} 
    In the encoder-decoder framework, there are two types of embeddings: one for numerical trajectories and another for symbolic functions. The goal is to align these two embedding spaces, enabling the system to directly generate the corresponding equation (\emph{i.e.}, the right-hand side of \Cref{eq:odeformer}) from input observational data of an unknown system.
    For the input observations, each number is represented as its own token. This tokenization scheme reduces the potentially infinite vocabulary size required to represent floating-point values to just $10,203$ tokens, including symbols such as $+$, $-$, digits ($0$ to $9$), and scientific notation components ($E-100$ to $E+100$). Each dimension's token sequence is fed separately into an embedding layer, and the resulting embeddings are concatenated to form a representation in $\mathbb{R}^{((n+1)\times 3)\times d_{\text{emb}}}$, where $d_{\text{emb}}$ is the dimension of the numerical trajectories embedding. 

    To encode mathematical expressions, the decoder's vocabulary includes tokens for all operators and variables, in addition to the tokens used for floating-point values. Expressions are represented in prefix notation, which eliminates the need for the model to predict parentheses. For example, the target sequence for the ODE $f(u) = \text{cos}(2.4242u)$ is represented as the six-token sequence: $[\rm{cos}~\rm{mul}~+~2424~\rm{E}-3~u ]$. 

    \textbullet~\textbf{Evaluation} ODEFormer introduces a large-scale dataset of 50M synthetically generated ODEs. These ODEs are randomly sampled with tree-based symbolic generation methods, incorporating unary and binary operators, constants, and variables. Generated trajectories are filtered for diversity by removing divergent or overly stable systems, ensuring a representative dataset.
	
    Unlike traditional symbolic regression methods that rely heavily on finite difference approximations, ODEFormer avoids direct derivative computation during training and instead learns directly from noisy, irregularly sampled solution trajectories, improving robustness and generalization.
	
	\textbullet~\textbf{Optimization.} ODEFormer uses a transformer-based encoder-decoder architecture with 86M parameters. It is trained end-to-end using cross-entropy loss to predict tokenized symbolic sequences. During inference, beam sampling with a size of 50 and temperature adjustments ensures diverse predictions. Optional post-hoc parameter optimization is performed using gradient-based methods to refine constants in the predicted ODEs.
	
\end{tcolorbox}

\end{document}